\PassOptionsToPackage{numbers}{natbib}
\PassOptionsToPackage{sort}{natbib}

\documentclass{article}

\usepackage{geometry}
\usepackage[utf8]{inputenc} 
\usepackage[T1]{fontenc}   
\usepackage{fullpage}
\usepackage[numbers]{natbib}

\date{}

\usepackage[utf8]{inputenc} 
\usepackage[T1]{fontenc}    
\usepackage{hyperref}       
\usepackage{url}            
\usepackage{booktabs}       
\usepackage{amsfonts}       
\usepackage{nicefrac}       
\usepackage{microtype}      
\usepackage{xcolor}         
\usepackage{amsmath}
\usepackage{subcaption}
\usepackage{algorithm}
\usepackage{algorithmic}
\usepackage{titletoc}
\usepackage{graphicx}
\usepackage{wrapfig}
\usepackage{cleveref}
\usepackage{booktabs}
\usepackage{colortbl}
\usepackage{xcolor}
\usepackage{multirow}
\usepackage{bm} 
\usepackage{caption}
\usepackage{bold-extra} 
\usepackage{listings}
\usepackage{multibib}
\newcites{appendix}{Additional References for Appendix}
\definecolor{lightgray}{gray}{0.9}

\usepackage{amsmath,amsfonts,bm}

\def\eqref#1{equation~\ref{#1}}

\def\1{\bm{1}}

\def\rpca{{\texttt{Robust-PCA }}}

\newcommand{\bA}{{\bf A}}

\newcommand{\bM}{{\bf M}}
\newcommand{\bL}{{\bf L}}
\newcommand{\bS}{{\bf S}}

\newcommand{\bW}{{\bf W}}
\newcommand{\bB}{{\bf B}}
\newcommand{\bz}{{\bf z}}

\DeclareMathAlphabet{\mathsfit}{\encodingdefault}{\sfdefault}{m}{sl}
\SetMathAlphabet{\mathsfit}{bold}{\encodingdefault}{\sfdefault}{bx}{n}

\title{FedRPCA: Enhancing Federated LoRA Aggregation Using Robust PCA}

\author{
\begin{tabular}{ccc}
\textbf{Divyansh Jhunjhunwala}\textsuperscript{1} & \textbf{Arian Raje}\textsuperscript{1} & \textbf{Madan Ravi Ganesh}\textsuperscript{2} \\ [3ex]
\textbf{Chaithanya Kumar Mummadi}\textsuperscript{2} & \textbf{Chaoqun Dong}\textsuperscript{2} & \textbf{Jiawei Zhou}\textsuperscript{3}\\ [3ex]
\textbf{Wan-Yi Lin}\textsuperscript{2} & \textbf{Gauri Joshi}\textsuperscript{1} & \textbf{Zhenzhen Li}\textsuperscript{2} \\
\end{tabular}
\\[12ex]
\textsuperscript{1}Carnegie Mellon University \quad
\textsuperscript{2}Bosch Center for AI \quad
\textsuperscript{3}Stony Brook University
}

\newif\ifdraft
\drafttrue        

\ifdraft
  \newcommand{\divyansh}[1]{{\leavevmode\color{orange}[DJ: #1]}}
  \newcommand{\baris}[1]{{\small\leavevmode\color{blue}[BA: #1]}}
  \newcommand{\GJ}[1]{{\leavevmode\color{magenta}[GJ: #1]}}
  \newcommand{\AR}[1]{{\leavevmode\color{red}[AR: #1]}}
  
  \definecolor{mybrown}{RGB}{165, 42, 42}
  \newcommand{\kumar}[1]{{\leavevmode\color{mybrown}[Kumar: #1]}}

  \newcommand{\jz}[1]{{\leavevmode\color{green}[JZ: #1]}}

\else
  \newcommand{\divyansh}[1]{}
  \newcommand{\baris}[1]{}
  \newcommand{\GJ}[1]{}
  \newcommand{\AR}[1]{}
  \newcommand\kumar[1]{}
  \newcommand{\jz}[1]{}
\fi

\begin{document}

\maketitle

\begin{abstract}
  \texttt{LoRA} has emerged as one of the most promising fine-tuning techniques, especially for federated learning (FL), since it significantly reduces communication and computation costs at resource-constrained clients. However, data heterogeneity remains a significant challenge for \texttt{LoRA}-based FL, and the conventional aggregation strategy based on \texttt{FedAvg} suffers from slow convergence and suboptimal accuracy. Motivated by recent advances in model merging, particularly \texttt{Task Arithmetic}, we explore the idea of aggregating client \texttt{LoRA} parameters using scaled averaging. We first observe that a naive application of \texttt{Task Arithmetic} is ineffective due to the high cosine similarity between client updates, indicating significant common knowledge in the updates across clients. To address this issue, we propose decomposing client \texttt{LoRA} updates via Robust Principal Component Analysis (\texttt{Robust-PCA}) into a common low-rank component and client-specific sparse components. Our proposed algorithm \texttt{FedRPCA} aggregates the low-rank components through averaging, consolidating common knowledge, and applies scaled averaging to the sparse components to amplify client-specific knowledge. We evaluate our approach across a variety of vision and language tasks and demonstrate that it achieves higher final accuracy and faster convergence compared to competing baselines.
\end{abstract}
\section{Introduction}
\label{sec:intro}
Pretrained models, including foundation models, have demonstrated strong generalization and zero-shot capabilities across domains such as vision, language, code, and scientific reasoning~\cite{radford2021learning, kirillov2023segment, oquab2023dinov2, touvron2023llama, yang2024qwen2, roziere2023code, guo2025deepseek, bommasani2021opportunities, brown2020language}. To extend their utility, there is a growing need to fine-tune these models on the vast and informative data available on edge devices such as smartphones. However, privacy and communication constraints make it impractical to transfer edge data to the cloud for centralized training. \emph{Federated Learning} (FL) offers a promising framework for collaboratively training models across decentralized data silos~\cite{mcmahan2017communication, kairouz2021advances}. To address the computational limitations of edge devices, \emph{parameter-efficient fine-tuning} (PEFT) techniques, especially \emph{low-rank adaptation} (\texttt{LoRA})~\cite{hu2022lora} has emerged as a practical and scalable solution. While \texttt{LoRA} significantly reduces compute and memory costs, it does not address the core challenge of data heterogeneity in FL. Under high heterogeneity, federated \texttt{LoRA} continues to suffer from slow convergence and degraded model quality \cite{singhal2024fedexlora, singhal2025fed}. This degradation can be thought of as primarily stemming from the \texttt{FedAvg} \cite{mcmahan2017communication} style averaging used to aggregate local \texttt{LoRA} parameters: although each client’s \texttt{LoRA} parameters can capture client-specific knowledge, vanilla averaging fails to preserve this knowledge in the resulting global \texttt{LoRA} parameters.

\begin{figure}[h]
\centering
   \subfloat[Original updates]
   {\includegraphics[width=0.33\linewidth]{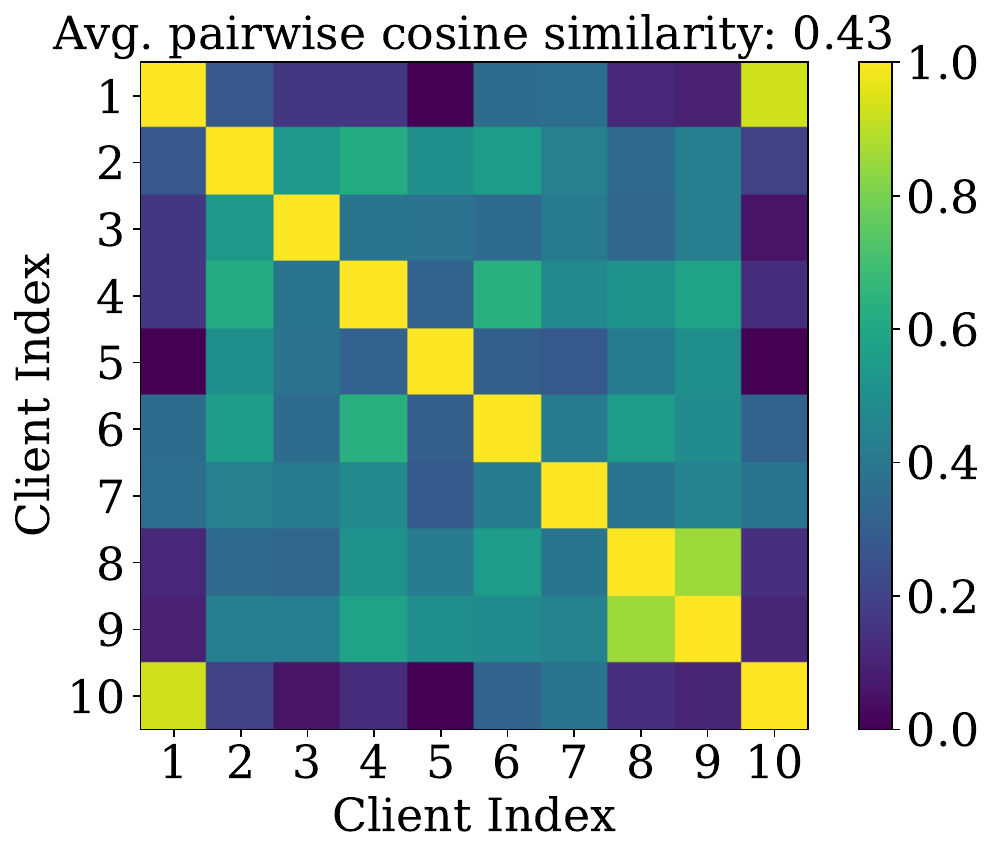}
    \label{subfig:cosine_sim_a}}
    \subfloat[Low-rank components]
   {\includegraphics[width=0.33\columnwidth]{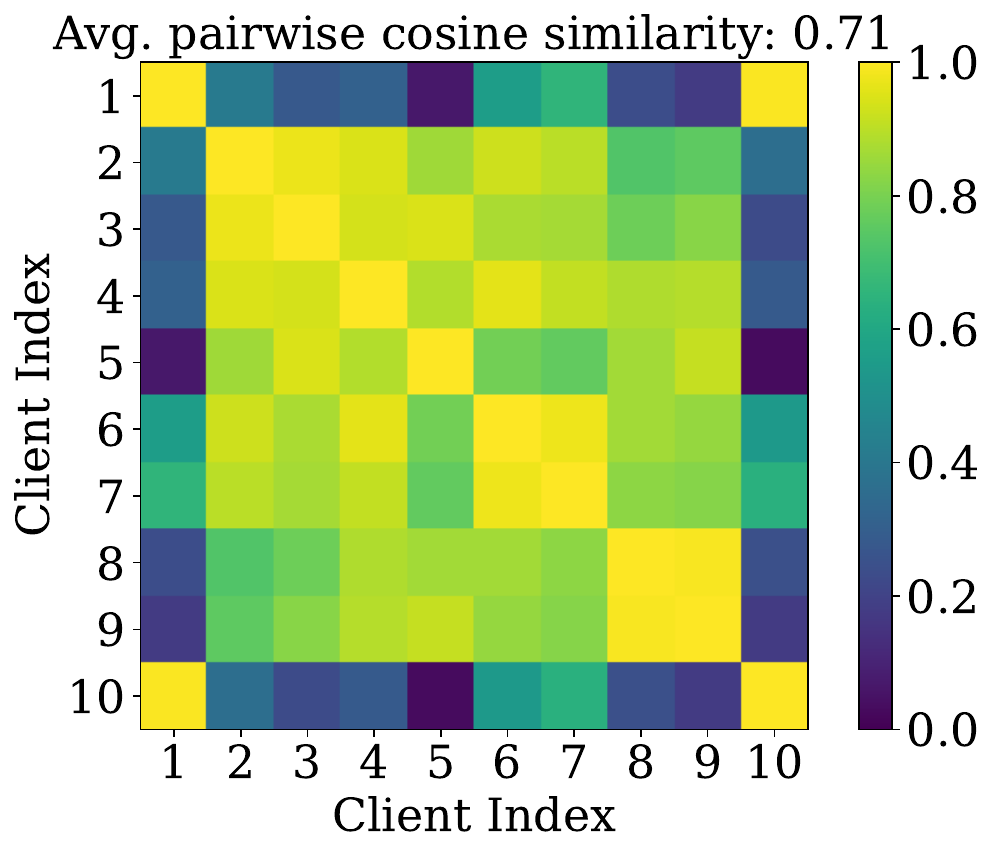}
    \label{subfig:cosine_sim_b}}
    \subfloat[Sparse components]
    {\includegraphics[width=0.33\columnwidth]{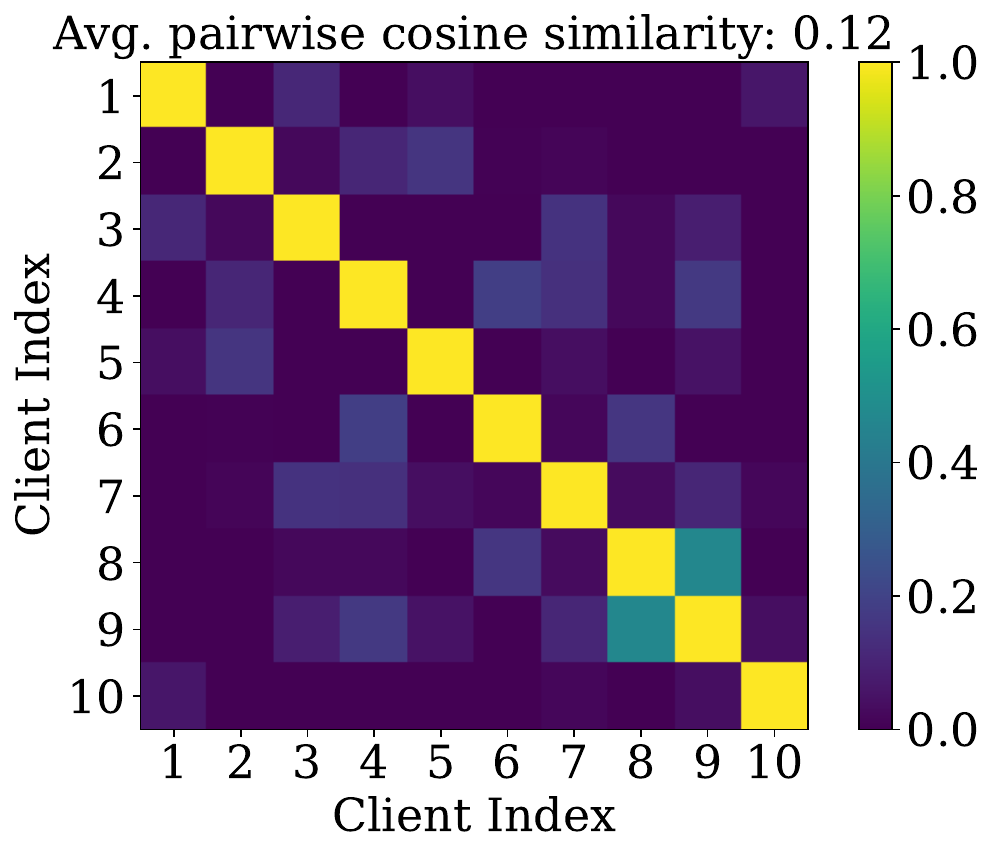}
    \label{subfig:cosine_sim_c}}
\captionsetup{font = small}
\caption{Cosine similarity matrices for the original client updates (\Cref{subfig:cosine_sim_a}), extracted low-rank components (\Cref{subfig:cosine_sim_b}), and sparse components (\Cref{subfig:cosine_sim_c}) corresponding in a federated \texttt{LoRA} setup (see Appendix for details). The average pairwise cosine similarity is substantially higher for the low-rank components, confirming that they capture shared structure, and significantly lower for the sparse components, confirming that they capture client-specific variations.}
\label{fig:correlation}
\end{figure}

\begin{wrapfigure}{r}{0.4\textwidth}
\centering
\vspace{-1.5em}
\includegraphics[width=0.4\textwidth]{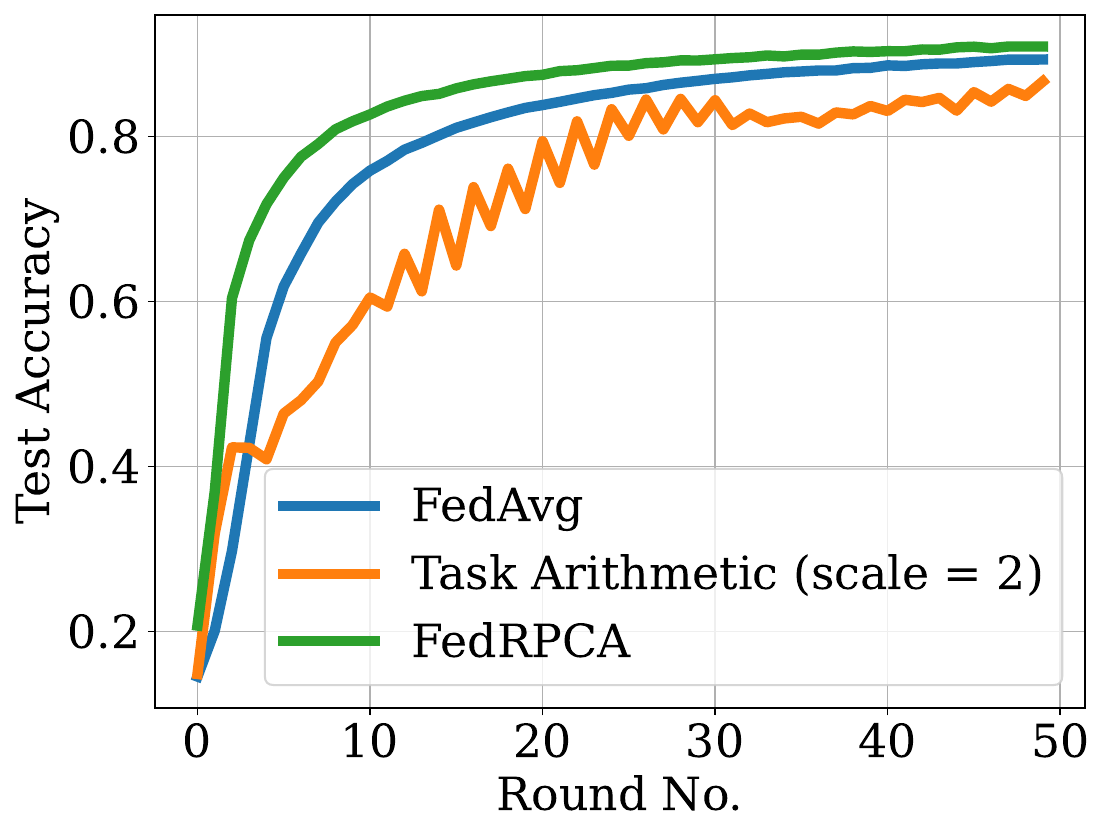} \vspace{-0.9em}
\captionsetup{font=small}
\vspace{-0.5em}
\captionsetup{font = small}
\caption{Performance comparison of \texttt{FedAvg}, \texttt{Task Arithmetic}, and \texttt{FedRPCA} on a federated \texttt{LoRA} task. A direct application of \texttt{Task Arithmetic} underperforms due to over-amplification of the common signal across clients. In contrast, \texttt{FedRPCA} decomposes client updates into common and client-specific components, selectively amplifying only the latter resulting in improved stability and performance.
}
\vspace{-1em}
\label{fig:intro_acc_plot}
\end{wrapfigure}
A promising approach to address this challenge is to reinterpret server-side aggregation in FL through the lens of \emph{model merging} \cite{MatenaR22}, where the goal is to combine multiple models fine-tuned on different tasks into a single, unified multi-task model. In particular, we focus on \texttt{Task Arithmetic}~\cite{IlharcoRWSHF23, tao2024task}, a widely-used model merging technique that represents each fine-tuned model as a \emph{task vector}—the difference between the fine-tuned model and the base model. Conceptually, the task vector captures the knowledge that must be added to the pre-trained base model to perform well on a given task. \texttt{Task Arithmetic} is based on the observation that task vectors corresponding to different tasks are often close to orthogonal, suggesting that the knowledge they encode is complementary. This motivates replacing naive averaging with \emph{scaled averaging} of task vectors, to better preserve knowledge from each individual task: $\bm{\tau}_{\text{merged}} = \beta \cdot \sum_{t \in \mathcal{T}} \bm{\tau}_t / |\mathcal{T}|$ where $\bm{\tau}$ represents a task vector, $\mathcal{T}$ is the set of tasks and $\beta > 1$ is a scaling factor. Given the success of \texttt{Task Arithmetic} in multi-task settings, a natural question arises: \emph{can Task Arithmetic-style scaled averaging be used to improve aggregation in federated LoRA fine-tuning by treating client updates as task vectors?}

Our results show that a direct application of \texttt{Task Arithmetic} is largely ineffective and in fact can perform worse than vanilla averaging (see \Cref{fig:intro_acc_plot}). We hypothesize that this stems from the fact that the knowledge in the client updates are not entirely complementary; \textit{they contain a substantial amount of common knowledge} as evidenced by the relatively high pairwise cosine similarity observed in the original \texttt{LoRA} updates (see \Cref{subfig:cosine_sim_a}). This similarity is expected in FL, where clients frequently have overlapping or related data distributions. As a result, a straightforward application of task arithmetic can end up amplifying the common knowledge in the global update, which is undesirable. To illustrate this issue, consider a toy example with two clients. Client 1 has data corresponding to a panda and a cat, while Client 2 has data from a panda and a dog. Let $\bm{\tau}_{P}$, $\bm{\tau}_{C}$, and $\bm{\tau}_{D}$ represent the signal or knowledge vectors corresponding to panda, cat, and dog, respectively. Following the \texttt{Task Arithmetic} view of additive composition, we can express the client updates as $\bm{\tau}_1 = \bm{\tau}_P + \bm{\tau}_C$ and $\bm{\tau}_2 = \bm{\tau}_P + \bm{\tau}_D$. The ideal global update would be $\bm{\tau}^* = \bm{\tau}_P + \bm{\tau}_C + \bm{\tau}_D$, which captures knowledge from all three classes. However, standard \texttt{FedAvg} produces the update $(\bm{\tau}_1 + \bm{\tau}_2)/2 = \bm{\tau}_P + \frac{1}{2}(\bm{\tau}_C + \bm{\tau}_D)$, under-representing the client-specific knowledge. On the other hand, \texttt{Task Arithmetic}-style scaled averaging with scaling factor $\beta = 2$ yields $\beta \cdot (\bm{\tau}_{1} + \bm{\tau}_{2})/2 = 2\bm{\tau}_P + \bm{\tau}_C + \bm{\tau}_D$, which over-amplifies the common signal $\bm{\tau}_P$.

Following this insight, we propose separating each client update into a \textit{common and client-specific component}. We achieve this using \emph{Robust Principal Component Analysis} (\texttt{Robust-PCA})~\cite{candes2011robust}, a technique which aims to decompose a given matrix $\bM$ as the sum of a low-rank matrix $\bL$ (capturing common structure among columns of $\bM$) and a sparse part $\bS$ (capturing localized variations in the columns of $\bM$), i.e. $\bM = \bL + \bS$. Pairwise cosine similarities of the extracted low-rank and sparse components for one of our federated \texttt{LoRA} setups are shown in \Cref{subfig:cosine_sim_b} and \Cref{subfig:cosine_sim_c}, respectively. Now assuming \texttt{Robust-PCA} correctly extracts the common knowledge in our example above, we obtain $\underbrace{[\bm{\tau}_1 \; \bm{\tau}_2]}_{\bM} \overset{\texttt{Robust-PCA}}{\rightarrow} \underbrace{[\bm{\tau}_P \; \bm{\tau}_P]}_{\bL} + \underbrace{[\bm{\tau}_C \; \bm{\tau}_D]}_{\bS}.$ Our proposed algorithm, \texttt{FedRPCA}, then performs standard averaging on the low-rank components and scaled averaging on the sparse components:
$\frac{\bm{\tau}_P + \bm{\tau}_P}{2} + \beta \cdot \frac{\bm{\tau}_C + \bm{\tau}_D}{2} = \bm{\tau}_P + \beta \cdot \frac{\bm{\tau}_C + \bm{\tau}_D}{2}$. Setting $\beta = 2$ exactly recovers the ideal update $\bm{\tau}^* = \bm{\tau}_P + \bm{\tau}_C + \bm{\tau}_D$ capturing knowledge about both the common and client-specific classes. While real setups are significantly complex than this setup; this example captures the core intuition behind our approach.

We note that applying \texttt{Robust-PCA} involves a singular value decomposition (SVD); however, this computation is relatively lightweight in the context of \texttt{LoRA}-based optimization, where the dimensionality of the update matrix $\bM$ is typically on the order of $\sim 10^3$ (see \Cref{appendix:compute_efficiency} for a detailed discussion on computational cost). Since \texttt{LoRA} has become a de facto standard for FL with foundation models, \texttt{FedRPCA} integrates seamlessly into existing workflows, requiring no changes to client-side training and serving as a simple, server-side drop-in replacement for \texttt{FedAvg}. Our main contributions can be summarized as follows.

\begin{itemize}
    \item We introduce \texttt{FedRPCA}, a server-side aggregation method that applies \texttt{Robust-PCA} to decompose client updates into common and client-specific components, enabling selective amplification of client-specific signals (see \Cref{sec:method}).

    \item We conduct experiments across a broad range of vision and language benchmarks, demonstrating that \texttt{FedRPCA} consistently outperforms strong heterogeneity-tackling baselines, boosting accuracy by $1\%$ on DTD and 20News (see \Cref{tab:accuracy-only}).

    \item Since \texttt{FedRPCA} modifies only the server-side aggregation step, we show that it can be combined with algorithms such as \texttt{FedProx} and \texttt{SCAFFOLD} which only modify client-level optimization thereby improving their performance (see \Cref{fig:rpca_combination}).

    \item Ablation studies show that the performance gains of \texttt{FedRPCA} grow with increasing client heterogeneity and number of clients, underscoring its effectiveness in challenging FL settings (see \Cref{tab:heterogeneity}, \Cref{tab:client-count}).
    
\end{itemize}

\section{Related Work}

\paragraph{Parameter Efficient Fine-Tuning in FL.} Recent efforts have adapted large language models (LLMs) to federated learning (FL) using parameter-efficient fine-tuning (PEFT) methods such as \texttt{LoRA}~\cite{hu2022lora, QLoRA, kuo2024federated, mao2025survey}, which minimize training costs while avoiding inference-time overhead by merging low-rank updates into the base model. \texttt{FedIT}~\cite{zhang2024towards} introduced \texttt{LoRA}-based instruction tuning in FL. However, separately averaging \texttt{LoRA} matrices $\bA$ and $\bB$ yields inaccurate updates, as the true model change is their product. To address this, \texttt{FedEx-LoRA}~\cite{singhal2024fedexlora}, \texttt{FFA-LoRA}~\cite{sun2024improving}, \texttt{LoRA-FAIR}~\cite{bian2025lorafairfederatedlorafinetuning} propose improved aggregation schemes that better preserve the bilinear update structure or correct misalignment across client updates. To support device heterogeneity, \texttt{HetLoRA}~\cite{cho-etal-2024-heterogeneous}, \texttt{FlexLoRA}~\cite{bai2024federated}, and \texttt{FLoRA}~\cite{wang2024flora} enable clients to fine-tune with varying \texttt{LoRA} ranks, using adaptive merging strategies such as rank projection or normalization. Hybrid approaches like \texttt{FedLPP}~\cite{jianhao-etal-2024-promoting} and \texttt{FedQLoRA}~\cite{hu2025fedqlora} combine \texttt{LoRA} with quantization or low-precision training to further reduce memory and communication costs in FL settings.

\paragraph{Model Merging.} Model merging aims to combine parameters of models fine-tuned on different tasks into a single multi-task model. Building on the idea of \textit{data-free} model merging using task vectors introduced by \texttt{Task Arithmetic} \cite{IlharcoRWSHF23}, \texttt{TIES-Merging}~\cite{TIES} trims task vectors and applies majority voting on parameter signs to resolve conflicts and mitigate task interference during merging.  \texttt{PCB-Merging}~\cite{du2024parameter} generalizes this further by modeling both inter-task and intra-task interference. Similarly, \texttt{DARE}~\cite{yu2024language} shows that many parameters can be pruned during merging if the remaining ones are rescaled appropriately. \texttt{EMR}~\cite{huang2024emrmerging} and \texttt{Consensus-Merging}~\cite{wang2024localizing} maintain lightweight task-specific masks alongside the merged model, enabling improved accuracy at inference time with minimal overhead. These methods demonstrate that model merging need not fully collapse task identity but can flexibly reconstruct it when needed.

Parallel to these data-free approaches, data-based merging methods use held-out validation data to guide the merging process. \texttt{FisherMerge}~\cite{matena2022merging} uses the Fisher Information matrix to weight parameter averaging, while \texttt{RegMean}~\cite{jin2023dataless} leverages the Gram matrix of intermediate representations. More expressive frameworks like \texttt{Surgery}~\cite{pmlr-v235-yang24t}, \texttt{ADAMerging}~\cite{yang2024adamerging}, and \texttt{WeMoE}~\cite{tang2024merging} introduce additional trainable components (e.g., adapters, merging coefficients, mixture-of-expert modules) to further mitigate interference during or after merging.

\paragraph{Robust Principal Component Analysis.} Robust Principal Component Analysis (\texttt{Robust-PCA})\cite{Robust-PCA} decomposes a matrix into a low-rank component $\bL$ and a sparse component $\bS$ via Principal Component Pursuit, which relaxes the problem using the nuclear norm of $\bL$ and the $\ell_1$ norm of $\bS$. Variants such as \texttt{Online-RPCA}~\cite{NIPS2013_8f121ce0} and \texttt{Learned-RPCA}~\cite{cai2021learned} extend \texttt{Robust-PCA} to streaming and learned non-convex settings, respectively. \texttt{Robust-PCA} has been used in contexts such as image processing and video surveillance~\cite{vaswani2018robust, bouwmans2018applications, bouwmans2014robust}, corrupted data recovery~\cite{NIPS2010_fe8c15fe}, ranking and collaborative filtering~\cite{Robust-PCA}. 

We discuss additional related work in \Cref{appendix:related_work}.
\section{Preliminaries}

We consider the FL problem of collaboratively optimizing the parameters of a shared model $\bW \in \mathbb{R}^{d \times l}$ across a population of $M$ clients. The goal is to minimize the global objective function $F(\bW)$, defined as the average of local objective functions $F_i(\bW)$ over all clients: \begin{align} F(\bW) := \frac{1}{M} \sum_{i=1}^M F_i(\bW), \label{eq:prob_form} \end{align} where each local objective  $F_i(\bW) := \frac{1}{|\mathcal{D}i|} \sum_{\bz_i \in \mathcal{D}_i} \ell(\bW, \bz_i)$ is the empirical risk over the local dataset $\mathcal{D}_i$ held by client $i$. Here, $\ell(\cdot, \cdot)$ denotes a loss function and each clients's dataset $\mathcal{D}_i$ is drawn from a client specific distribution, potentially leading to heterogeneity across clients.  

\paragraph{LoRA Fine-Tuning.} When optimizing \( \bW \) starting from a pre-trained model \( \bW_0 \), \cite{hu2022lora} observes that the weight update \( \Delta \bW = \bW - \bW_0 \) is often low-rank. Building on this insight, the authors propose \texttt{LoRA}, a PEFT method that models the update \( \Delta \bW \) as the product of two low-rank matrices: \( \Delta \bW = \bB \cdot \bA \), where \( \bB \in \mathbb{R}^{d \times r} \), \( \bA \in \mathbb{R}^{r \times l} \) and \( r \ll d, l \).

\begin{align}
\underbrace{\bW}_{\text{Fine-tuned Weights}} \approx 
\underbrace{\bW_0}_{\text{Pretrained Weights (Frozen)}} + 
\underbrace{\bB \cdot \bA}_{\text{\texttt{LoRA} Weights (Trainable)}}
\end{align}

Note that the total number of trainable parameters under this formulation is \( r \times (d + l) \), which represents a substantial reduction compared to full-rank fine-tuning, which requires \( d \times l \) trainable parameters.

\paragraph{FedAvg with LoRA Fine-Tuning.} 
We adopt a similar setting as \cite{cho-etal-2024-heterogeneous, zhang2024towards} for extending \texttt{LoRA} fine-tuning to FL. We assume that all clients begin with access to a shared pre-trained model \( \bW_0 \). During each communication round \( t \) of the \texttt{FedAvg} algorithm, the central server broadcasts the current global \texttt{LoRA} weights \( \bA^{(t)} \) and \( \bB^{(t)} \) to all clients. Each client then performs \( \tau \) steps of local optimization (e.g., using SGD) on its local dataset to compute updated local \texttt{LoRA} weights \( \bA_i^{(t)} \) and \( \bB_i^{(t)} \). After local training, each client computes its \texttt{LoRA} updates relative to the global model:

\begin{align}
    \Delta \bB_i^{(t)} \leftarrow \bB_i^{(t)} - \bB^{(t)}, \quad \Delta \bA_i^{(t)} \leftarrow \bA_i^{(t)} - \bA^{(t)}.
\end{align}
These updates \( \Delta \bA_i^{(t)} \) and \( \Delta \bB_i^{(t)} \) are then sent back to the server, which performs a simple average to update the global \texttt{LoRA} parameters: 
\begin{align}
   \bB^{(t+1)} = \bB^{(t)} + \frac{1}{M} \sum_{i = 1}^M \Delta \bB_i^{(t)}, \quad \bA^{(t+1)} = \bA^{(t)} + \frac{1}{M} \sum_{i = 1}^M \Delta \bA_i^{(t)} 
\end{align}
This procedure enables communication-efficient and scalable fine-tuning of large pre-trained models in federated settings, leveraging the low-rank structure of \texttt{LoRA} while preserving client data privacy. We note that a recent line of work \cite{wang2024flora, singhal2024fedexlora, singhal2025fed} has proposed \textit{exact aggregation} schemes that aim to update the global model using the average of reconstructed local updates, \( \sum_{i=1}^M (\bB_i \cdot \bA_i) \), rather than aggregating \( \bB \) and \( \bA \) separately. These approaches are orthogonal to ours and can be easily incorporated into our framework as complementary enhancements.

\section{Motivation and Proposed Method}
\label{sec:method}

While the procedure above provides a simple and scalable way to fine-tune large pre-trained models, prior work has shown that in the presence of data heterogeneity, averaging-based approaches can suffer from slow convergence and degraded performance \cite{singhal2024fedexlora, singhal2025fed}. Improving upon vanilla averaging has therefore become a fundamental challenge in federated \texttt{LoRA} fine-tuning.

\paragraph{Scaled Averaging / Task Arithmetic.} One of the simplest techniques to improve upon vanilla averaging is to scale the averaged client updates when updating the global model:
\begin{align}
\label{eq:scaled_average}
\bB^{(t+1)} = \bB^{(t)} + \beta \cdot \frac{1}{M} \sum_{i = 1}^M \Delta \bB_i^{(t)}, \quad
\bA^{(t+1)} = \bA^{(t)} + \beta \cdot \frac{1}{M} \sum_{i = 1}^M \Delta \bA_i^{(t)},
\end{align}
where $\beta > 1$ is the scaling factor. In the FL literature, this idea has been proposed as a generalization of \texttt{FedAvg} that incorporates a \textit{server-side learning rate}\cite{ReddiCZGRKKM21, Jhunjhunwala0J23}. As discussed in \Cref{sec:intro}, a similar notion emerges in the model merging literature through \texttt{Task Arithmetic}\cite{IlharcoRWSHF23}, which observes that for full-model fine-tuning, the resulting \textit{task vectors} i.e., the difference between fine-tuned and pretrained weights tend to be nearly orthogonal across tasks. This low cosine similarity implies that the task vectors capture largely distinct knowledge. In such cases, scaled averaging helps amplify task-specific signals in each update, which would otherwise be diminished by uniform averaging. Empirically, \texttt{Task Arithmetic} demonstrates that using a scaling factor $\beta \in {2, 3}$ can significantly outperform vanilla averaging ($\beta = 1$) when merging independently fine-tuned models (see Fig. 15 in~\cite{IlharcoRWSHF23}). However, as shown in \Cref{fig:intro_acc_plot}, a direct application of this idea in FL proves largely ineffective due to the higher cosine similarity across client updates.

\paragraph{Separating Common Signal from Client-Specific Signal with Robust PCA.} 

\begin{wrapfigure}{r}{0.45\textwidth}
\centering 
\vspace{-1.5em}
\includegraphics[width=0.45\textwidth]{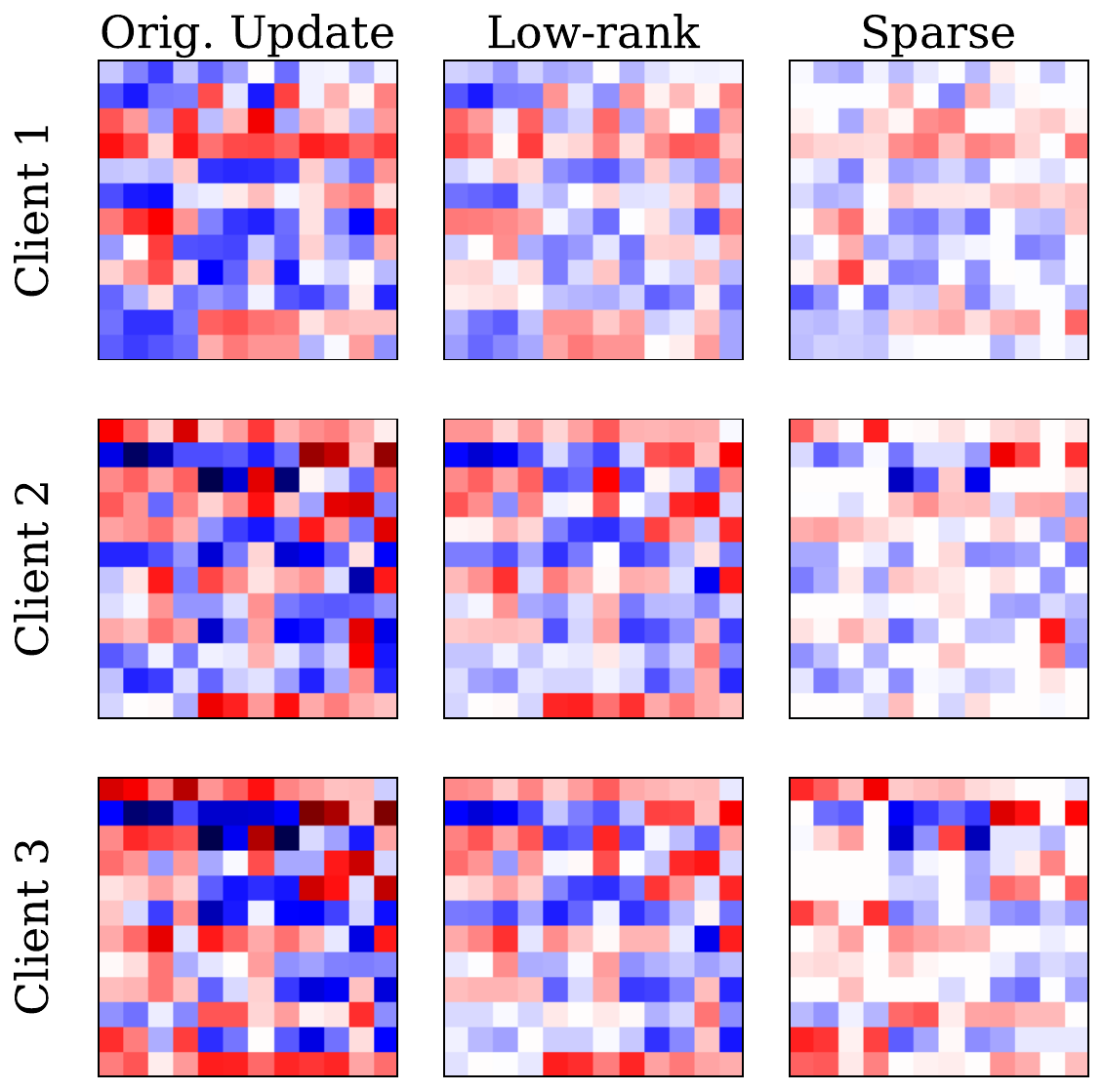}
\captionsetup{font=small}
\caption{Visualization of a subset of the original updates (left column), low-rank component (middle column) and sparse component (right column) for a federated \texttt{LoRA} setup. The low-rank components are similar across clients, while the sparse ones are distinct showing that \texttt{Robust-PCA} effectively isolates common versus client-specific signals.}
\vspace{-2em} 
\label{fig:lora_visualization}
\end{wrapfigure}
From the discussion above, it is evident that the effectiveness of scaled averaging depends on the low cosine similarity across the updates being merged. As such making scaled averaging effective for federated \texttt{LoRA} fine-tuning is non-trivial, as client updates contain a significant amount of common signal – which we do not want to amplify. This leads to the natural question \textit{can we extract the common signal from the client updates, leaving behind only the client-specific signal?} If so, we could apply scaling selectively on the client-specific signal, which would presumably have lower cosine similarity  and thereby avoid the instability which comes from amplifying the common signal.

We formalize this extraction as a \textbf{signal separation problem}, where the objective is to isolate a common signal from a set of noisy measurements. In our setting, the “noise” corresponds to client-specific signals, which we model as sparse motivated by findings from the model merging literature showing that task-specific updates are often highly sparse \cite{TIES}. A canonical method for this setting is \textbf{Robust Principal Component Analysis (\texttt{Robust-PCA})} \cite{Robust-PCA}, which decomposes a given matrix \( \bM \) into the sum of a low-rank matrix \( \bL \) and a sparse matrix \( \bS \):
\begin{align}
\bM = \underbrace{\bL}_{\text{Low-rank matrix}} + \underbrace{\bS}_{\text{Sparse matrix}}.
\end{align}
In this decomposition, the columns in the low-rank matrix \( \bL \) captures the common signal across clients, while the columns in sparse matrix \( \bS \) captures the client-specific deviations. \Cref{fig:lora_visualization} shows an example of the original client \texttt{LoRA} updates and the extracted low-rank and sparse components with \texttt{Robust-PCA}. With this formulation in place, we are now ready to present our proposed algorithm.

\setlength{\textfloatsep}{1em}
\begin{algorithm}[t]
\caption{\texttt{FedRPCA}} \label{algo1}
\begin{algorithmic}[1]
\label{alg:alg1}
\STATE \textbf{Input:} Pretrained model $\bW^{(0)}$, number of rounds $T$, local steps $K$, scaling factor $\beta$
\FOR{$t = 0$ \TO $T-1$}
    \STATE \textbf{Server broadcasts} current \texttt{LoRA} weights $\bA^{(t)}, \bB^{(t)}$ to all clients
    \FORALL{clients $i \in [M]$ \textbf{in parallel}}
        \STATE $\bA_i^{(t)}, \bB_i^{(t)} \leftarrow \texttt{Local-Optimization}(\bW^{(0)}, \bA^{(t)}, \bB^{(t)}, \mathcal{D}_i)$
        \STATE $\Delta \bA_i^{(t)} \leftarrow \bA_i^{(t)} - \bA^{(t)}$, $\Delta \bB_i^{(t)} \leftarrow  \bB_i^{(t)} - \bB^{(t)}$
        \STATE \textbf{Client $i$ sends} $\Delta \bA_i^{(t)}, \Delta \bB_i^{(t)}$ to server
    \ENDFOR
    \STATE \textbf{Server constructs} matrices $\bM_{\bA}^{(t)}$, $\bM_{\bB}^{(t)}$ by stacking vectorized client updates (Eqs.~\ref{eq:M_A}, \ref{eq:M_B})
    \STATE \textbf{Apply RPCA:}
    $$
    \bL_{\bA}, \bS_{\bA} \leftarrow \texttt{Robust-PCA}(\bM_{\bA}^{(t)}), \quad \bL_{\bB}, \bS_{\bB} \leftarrow \texttt{Robust-PCA}(\bM_{\bB}^{(t)})
    $$
    \STATE \textbf{Update \texttt{LoRA} weights:}
    \begin{align*}
        \bA^{(t+1)} &= \bA^{(t)} + \mathrm{reshape}\left(\frac{1}{M} \sum_{i=1}^M (\bL_{\bA})_{:, i} \right) + \beta \cdot \mathrm{reshape}\left(\frac{1}{M} \sum_{i=1}^M (\bS_{\bA})_{:, i} \right) \\
        \bB^{(t+1)} &= \bB^{(t)} + \mathrm{reshape}\left(\frac{1}{M} \sum_{i=1}^M (\bL_{\bB})_{:, i} \right) + \beta \cdot \mathrm{reshape}\left(\frac{1}{M} \sum_{i=1}^M (\bS_{\bB})_{:, i} \right)
    \end{align*}
\ENDFOR
\end{algorithmic}
\end{algorithm}

\paragraph{Proposed Algorithm.} Let \( \{\Delta \bA_i^{(t)}\}_{i=1}^M \) and \( \{\Delta \bB_i^{(t)}\}_{i=1}^M \) denote the \texttt{LoRA} updates collected from \( M \) clients at round \( t \). We construct matrices \( \bM_{\bA}^{(t)} \) and \( \bM_{\bB}^{(t)} \) by vectorizing each client’s update and stacking them column-wise:
\begin{align}
   \bM_{\bA}^{(t)} &= [\mathrm{vec}(\Delta \bA_1^{(t)}) \; \mathrm{vec}(\Delta \bA_2^{(t)}) \; \dots \; \mathrm{vec}(\Delta \bA_M^{(t)})] \in \mathbb{R}^{(r \cdot l) \times M}, \label{eq:M_A} \\
   \bM_{\bB}^{(t)} &= [\mathrm{vec}(\Delta \bB_1^{(t)}) \; \mathrm{vec}(\Delta \bB_2^{(t)}) \; \dots \; \mathrm{vec}(\Delta \bB_M^{(t)})] \in \mathbb{R}^{(r \cdot d) \times M}, \label{eq:M_B}
\end{align}
where \( \mathrm{vec}(\cdot) \) denotes the vectorization operation that flattens a matrix into a column vector.

We apply \rpca independently to both \( \bM_{\bA}^{(t)} \) and \( \bM_{\bB}^{(t)} \) to obtain their respective common and client-specific decompositions:
\begin{align}
    \bL_{\bA}, \bS_{\bA} &\leftarrow \texttt{Robust-PCA}(\bM_{\bA}^{(t)}) \hspace{5pt} , \hspace{5pt}
    \bL_{\bB}, \bS_{\bB} \leftarrow \texttt{Robust-PCA}(\bM_{\bB}^{(t)}),
\end{align}
where \( \bL_{\bA}, \bL_{\bB} \) are low-rank matrices capturing common signal across clients, and \( \bS_{\bA}, \bS_{\bB} \) represent sparser, client-specific deviations.

Now given this decomposition, to construct the merged update, we average across the columns of each component and apply a scaling factor \( \beta \) only to the client-specific infor:
\begin{align}
    \bA^{(t+1)} &= \bA^{(t)} + \mathrm{reshape}\left(\frac{1}{M} \sum_{i=1}^M (\bL_{\bA})_{:, i} \right) + \beta \cdot \mathrm{reshape}\left( \frac{1}{M} \sum_{i=1}^M (\bS_{\bA})_{:, i} \right), \\
    \bB^{(t+1)} &= \bB^{(t)} + \mathrm{reshape}\left(\frac{1}{M} \sum_{i=1}^M (\bL_{\bB})_{:, i} \right) + \beta \cdot \mathrm{reshape}\left( \frac{1}{M} \sum_{i=1}^M (\bS_{\bB})_{:, i} \right),
\end{align}
where \( \mathrm{reshape}(\cdot) \) restores the vectorized updates to their original matrix dimensions. We refer to our proposed method as \textit{\underline{F}ederated \underline{R}obust \underline{PCA} Decomposed Aggregation} (\texttt{FedRPCA}), after the \texttt{Robust-PCA} procedure that forms the core of the algorithm. An algorithmic outline of \texttt{FedRPCA} is presented in \Cref{alg:alg1}.

\paragraph{Choice of $\beta$.} 
\begin{wrapfigure}{r}{0.35\textwidth}
\vspace{-1.5em}
\centering 
\includegraphics[width=0.35\textwidth]{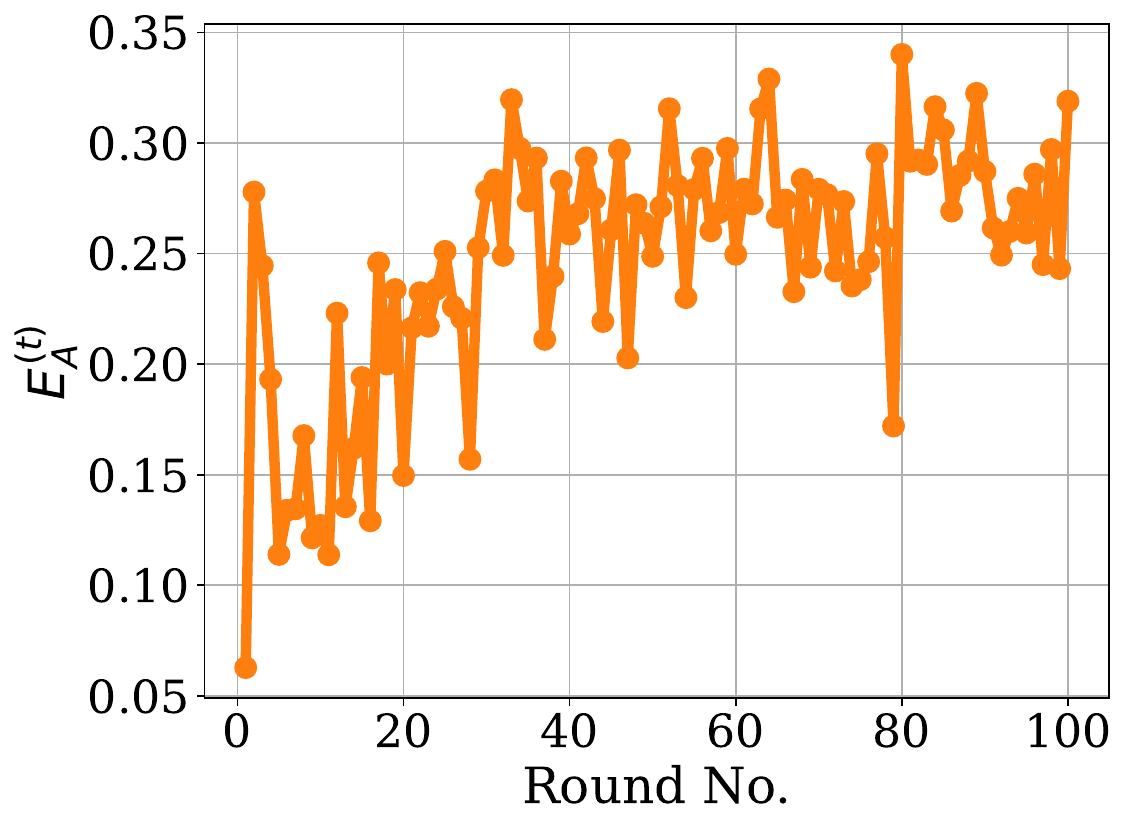} 
\captionsetup{font=small}
\vspace{-2.2em}
\captionsetup{font = small}
\caption{Evolution of $E_{\bA}^{(t)}$, which quantifies the relative contribution of the client-specific component to the overall update across training rounds. The increasing trend in later rounds highlights the need for an adaptive choice of $\beta$ for best performance.}
\vspace{-6em}
\label{fig:adapt_beta}
\end{wrapfigure}
To set $\beta$, we find it important to consider the relative contribution of sparse components to the overall update. We quantify this using the ratio of the norm of the sum of the sparse components to the norm of the sum of the original updates: $E_{\bA}^{(t)} = \frac{\| \bS_{\bA}^{(t)} \cdot \bm{1} \|}{\| \bM_{\bA}^{(t)} \cdot \bm{1}  \|}$ (with a similar definition for $\bB$). If this ratio is small, a higher value of $\beta$ works well while a smaller value works well when this ratio is large. As shown in \Cref{fig:adapt_beta} this ratio varies throughout training, typically small in early rounds when the common signal dominates, and larger later as client-specific signals grow. This motivates using a dynamic $\beta$ rather than a fixed one and we adopt a simple heuristic: set $\beta^{(t)}_{\bA} = 1 / E_{\bA}^{(t)}, \beta^{(t)}_{\bB} = 1 / E_{\bB}^{(t)}$ to reflect this. We find this adaptive choice generally outperforms constant scaling. Further details are provided in the Appendix.

\section{Experiments}
\label{sec:expts}

Our objective in this section is to evaluate whether \texttt{FedRPCA} can consistently enhance the performance of \texttt{LoRA} fine-tuning in federated settings. To that end, we conduct experiments across a diverse set of vision and language tasks, comparing our method against several competitive baselines that have been proposed to mitigate the challenges posed by client heterogeneity. 
\vspace{-0.8em} 
\paragraph{Baselines.}  
We group our baselines into two main categories:  
\textbf{(1) Client-level optimization methods}, which modify the local training objective or update rule to address heterogeneity; and  
\textbf{(2) Server-level aggregation methods}, which focus on improving how the server aggregates client updates. In the first category, we consider \bm{$(i)$} \texttt{SCAFFOLD} \cite{KarimireddyKMRS20}, which uses control variates to reduce the variance introduced by client drift during local updates \bm{$(ii)$} \texttt{FedProx} \cite{FedProx}, which adds a proximal term to the local objective to discourage divergence from the global model \bm{$(iii)$} \texttt{MOON} \cite{MOON}, which introduces contrastive learning between local and global models to enforce consistency during local training. In the second category, we compare against: \bm{$(i)$} \texttt{Task Arithmetic} \cite{IlharcoRWSHF23}, which performs the scaled averaging based merging operation described in \Cref{eq:scaled_average} \bm{$(ii)$} \texttt{TIES-Merging} \cite{TIES} which seeks to improve model merging by explicitly addressing parameter interference, selectively resetting small updates, resolving parameter-sign conflicts, and merging only aligned parameters.
\vspace{-0.8em}
\paragraph{Datasets, Models and Experiment Setup.}  

For our vision task experiments, we evaluate our method on four standard image classification benchmarks:  
\bm{$(i)$} \textbf{SVHN}~\cite{SVHN}, a digit classification dataset with over 73,000 training and 26,000 test images across 10 classes; \bm{$(ii)$} \textbf{DTD}~\cite{DTD}, a texture classification dataset consisting of 4,000 images spanning 47 describable texture classes; \bm{$(iii)$} \textbf{EuroSAT}~\cite{EuroSAT}, a satellite image classification dataset consisting of 27,000 images across 10 classes and  \bm{$(iv)$} \textbf{Stanford Cars}~\cite{Stanford_Cars}, a fine-grained classification dataset of 16,185 car images from 196 categories. To simulate non-IID client distributions, we partition each dataset across 50 clients using a Dirichlet distribution with concentration parameter $\alpha = 0.3$, following standard practice~\cite{Dirichlet-NonIID}. For all image tasks, we use a \textbf{CLIP ViT-B/32} backbone \cite{ClipVit} and report final test accuracy.

For our language model experiments, we evaluate on two tasks: \bm{$(i)$} \textbf{20News}~\cite{20NewsGroup}, a document classification benchmark containing approximately 11,000 news articles across 20 categories; and  
\bm{$(ii)$} \textbf{MRQA}~\cite{MRQA}, a machine reading comprehension benchmark that unifies several extractive QA datasets. We use \textbf{GPT-2}~\cite{gpt2} for 20News and \textbf{T5-Base}~\cite{T5} for MRQA. Accuracy is used as the evaluation metric for 20News, while MRQA is evaluated using token-level F1. For 20News, we simulate heterogeneity using a Dirichlet partition with \( \alpha = 0.3 \); for MRQA, we treat the six MRQA subtasks as labels and partition examples using Dirichlet heterogeneity in a similar fashion.

In all experiments, we freeze the pre-trained backbone and apply \texttt{LoRA} fine-tuning with rank $r = 4$ only to the Query and Value projection matrices in the self-attention layers. All $50$ clients participate in each communication round, performing 1 local epochs before aggregation. Fine-tuning is performed using the \texttt{Adam} optimizer with a learning rate of $1 \times 10^{-4}$. For EuroSAT and SVHN we perform $100$ rounds of fine-tuning and $200$ rounds for the rest. Additional experimental details and hyperparameter tuning procedures are provided in \Cref{appendix:hyperparameters}.

\begin{table}[t]
\setlength{\tabcolsep}{4pt}
\centering
\footnotesize
\captionsetup{font = small}
\caption{We report final accuracy for all datasets, except MRQA, where we use token-level F1 score. The bottom row shows the improvement of \texttt{FedRPCA} over the second-best method. \texttt{FedRPCA} consistently outperforms all baselines across datasets.}
\label{tab:accuracy-only}
\begin{tabular}{lcccccc}
\toprule
\textbf{Method} & \textbf{EuroSAT} & \textbf{SVHN} & \textbf{DTD} & \textbf{Cars} & \textbf{20News} & \textbf{MRQA}\\
\midrule
\texttt{FedAvg}  & $96.41$\tiny{$\pm 0.10$} & $91.57$\tiny{$\pm 0.06$} & $51.99$\tiny{$\pm 0.03$} & $62.21$\tiny{$\pm 0.05$}  & $61.73$\tiny{$\pm 0.14$} & $62.50$\tiny{$\pm 0.17$} \\
\addlinespace
\texttt{SCAFFOLD}  & $96.13$\tiny{$\pm 0.04$} & $89.11$\tiny{$\pm 0.12$} & $39.64$\tiny{$\pm 1.90$} & $59.96$\tiny{$\pm 0.13$}  & $52.93$\tiny{$\pm 0.43$} & $59.64$\tiny{$\pm 0.11$} \\
\addlinespace
\texttt{FedProx}   & $96.41$\tiny{$\pm 0.10$} & $91.56$\tiny{$\pm 0.07$} & $51.99$\tiny{$\pm 0.06$} & $62.22$\tiny{$\pm 0.05$}  & $61.73$\tiny{$\pm 0.09$} & $62.62$\tiny{$\pm 0.19$} \\
\addlinespace 
\texttt{MOON}  & $96.42$\tiny{$\pm 0.09$} & $91.57$\tiny{$\pm 0.06$} & $52.03$\tiny{$\pm 0.08$} & $62.22$\tiny{$\pm 0.05$}  & $62.19$\tiny{$\pm 0.50$} & $62.77$\tiny{$\pm 0.87$}\\
\addlinespace
\texttt{Task Arithmetic}  & $96.58$\tiny{$\pm 0.50$} & $91.49$\tiny{$\pm 0.58$} & $53.87$\tiny{$\pm 0.55$} & $62.60$\tiny{$\pm 0.17$}  & $62.79$\tiny{$\pm 1.03$} & $63.16$\tiny{$\pm 0.62$} \\
\addlinespace
\texttt{TIES-Merging}   & $95.76$\tiny{$\pm 0.30$} & $89.51$\tiny{$\pm 1.10$} & $54.17$\tiny{$\pm 0.48$} & $62.07$\tiny{$\pm 0.14$}  & $61.96$\tiny{$\pm 0.97$} & $63.11$\tiny{$\pm 0.78$} \\
\addlinespace
\rowcolor{lightgray}
\texttt{FedRPCA} (Ours)  & \bm{$97.15$}\tiny{$\pm 0.04$} & \bm{$92.40$}\tiny{$\pm 0.11$} & \bm{$55.18$}\tiny{$\pm 0.45$} & \bm{$63.45$}\tiny{$\pm 0.04$}  & \bm{$63.78$}\tiny{$\pm 0.17$} & \bm{$63.44$}\tiny{$\pm 0.11$} \\
\midrule
\textit{Improvement} & \(+\mathbf{0.57}\) & \(+\mathbf{0.83}\) & \(+\mathbf{1.01}\) & \(+\mathbf{0.85}\) & \(+\mathbf{0.99}\) & \(+\mathbf{0.28}\) \\
\bottomrule
\end{tabular}
\label{table:main_result}
\vspace{0.5em}
\end{table}

\paragraph{Takeaways.} From \Cref{table:main_result}, we see that \texttt{FedRPCA} consistently outperforms all baselines, achieving up to a $1.01\%$ improvement in final accuracy, particularly on more challenging datasets such as DTD and Stanford Cars. We also observe that methods which solely modify local optimization such as \texttt{SCAFFOLD} and \texttt{FedProx} often perform comparably to or even worse than \texttt{FedAvg}. This phenomenon has also been reported in prior work \cite{Chen0LSC23, NguyenWMSR23} when doing full model fine-tuning in FL and is commonly attributed to the use of a strong pretrained initialization, which diminishes the relative impact of client-side correction methods. Our results show that this phenomenon extends to federated \texttt{LoRA} fine-tuning as well. These results underscore the importance of effective server-level aggregation strategies for addressing heterogeneity in such settings. At the same time, the gains for methods that directly adopt model merging techniques such as \texttt{Task Arithmetic} and \texttt{TIES-Merging} are also found to be limited, particularly for vision tasks. As previously discussed, \texttt{Task Arithmetic} assumes low cosine similarity among task vectors, an assumption that typically does not hold in our federated setting. Similarly, we hypothesize that \texttt{TIES-Merging}, which relies on sign-based alignment, is sensitive to such high similarity and fails to deliver meaningful improvements when this assumption is violated.

\subsection{Ablation Experiments}

In this subsection, we explore additional properties of \texttt{FedRPCA}, including its compatibility with local optimization methods and the impact of rank, data heterogeneity, and the number of clients on its performance. Unless stated otherwise, all experiments use the SVHN dataset with 50 clients and a heterogeneity level of 0.3. Ablation results on 20News dataset is provided in \Cref{appendix:ablation_20News}.

\vspace{-1.5em}
\paragraph{Combining RPCA with Local Optimization.}
\begin{wrapfigure}{r}{0.37\textwidth}
\centering
\vspace{-2.0em}
\includegraphics[width=0.33\textwidth]{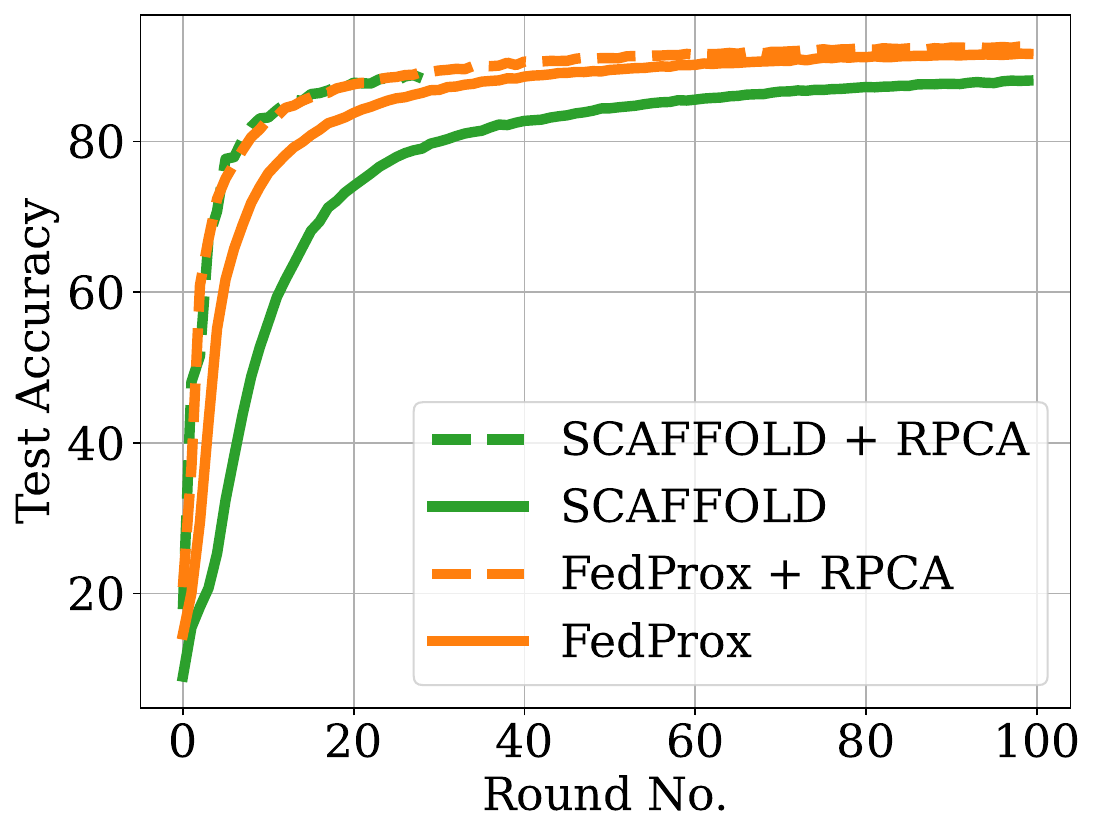}
\captionsetup{font = small} 
\vspace{-1.0em}
\caption{Performance for \texttt{FedRPCA} combined with \texttt{FedProx} and \texttt{SCAFFOLD}.}
\vspace{-4em}
\label{fig:rpca_combination}
\end{wrapfigure}
One of the key strengths of \texttt{FedRPCA} is its modularity–it operates entirely at the server level without making assumptions about the client-side training process. This design allows it to be seamlessly combined with client-level optimization strategies such as \texttt{FedProx} (which penalizes local model divergence) and \texttt{SCAFFOLD} (which mitigates client drift using control variates). \Cref{fig:rpca_combination} presents the results of these combinations, showing that \texttt{FedRPCA} can be successfully integrated with these methods, highlighting their complementary nature.
\vspace{-0.5em}
\paragraph{Impact of Heterogeneity.}
To simulate varying degrees of data heterogeneity, we adjust the Dirichlet concentration parameter $\alpha$ during data partitioning. Lower values of $\alpha$ yield more skewed partitions, where individual clients observe only a smaller subset of the overall label space. The results, summarized in \Cref{tab:heterogeneity}, show that \textbf{\texttt{FedRPCA} offers increasing improvements over baselines as heterogeneity grows}. This is expected, as greater heterogeneity introduces more client-specific signal in the updates, which naive averaging tends to suppress. Thus, in such settings, \texttt{FedRPCA}'s ability to separate and selectively amplify client-specific signals becomes increasingly important, highlighting its effectiveness against non-IID data distributions.
\vspace{-1em}
\paragraph{Impact of Number of Clients.}
We study how performance varies with the number of clients in our FL setup in \Cref{tab:client-count}. We observe that \textbf{the performance gap between \texttt{FedRPCA} and baseline methods widens as the number of clients increases}. This trend can be attributed to two factors: the increased dampening effect introduced by naive averaging over a larger set of client updates, and the growing need to preserve and amplify client-specific signals that may otherwise be lost. This observation is consistent with findings in \texttt{Task Arithmetic}, where the benefits of scaled averaging become more pronounced as the number of merged task vectors increases.

\noindent
\begin{minipage}[t]{0.48\textwidth}
\centering
\renewcommand{\arraystretch}{1.32}
\setlength{\tabcolsep}{0.1pt}
\footnotesize
\captionsetup{font = small}
\captionof{table}{Final accuracy (\%) for varying heterogeneity (a lower $\alpha$ indicates more heterogeneity). Improvement offered by \texttt{FedRPCA} increases as heterogeneity increases. \texttt{TA} abbreviates \texttt{Task Arithmetic}.}
\label{tab:heterogeneity}
\begin{tabular}{lccc}
\toprule
\textbf{Method} & \bm{$\alpha = 10$} & \bm{$\alpha = 1.0$} & \bm{$\alpha = 0.1$} \\
\midrule
\texttt{FedAvg}       & $91.78$\tiny{$\pm 0.05$} & $91.54$\tiny{$\pm 0.06$} & $89.46$\tiny{$\pm 0.14$} \\
\texttt{FedProx}      & $91.77$\tiny{$\pm 0.04$} & $91.54$\tiny{$\pm 0.06$} & $89.44$\tiny{$\pm 0.11$} \\
\texttt{SCAFFOLD} & $91.11$\tiny{$\pm 0.08$} & $90.49$\tiny{$\pm 0.07$} & $83.33$\tiny{$\pm 0.42$} \\
\texttt{MOON} & $91.76$\tiny{$\pm 0.07$} & $91.51$\tiny{$\pm 0.05$} & $89.53$\tiny{$\pm 0.08$} \\
\texttt{TIES-Merging} & $88.89$\tiny{$\pm 1.01$} & $90.41$\tiny{$\pm 0.42$} & $83.20$\tiny{$\pm 5.47$} \\
\texttt{TA} & $91.57$\tiny{$\pm 0.31$} & $90.22$\tiny{$\pm 0.48$} & $88.56$\tiny{$\pm 0.84$} \\
\rowcolor{lightgray}
\texttt{FedRPCA} & \bm{$92.80$}\tiny{$\pm 0.05$} & \bm{$92.50$}\tiny{$\pm 0.03$} & \bm{$91.33$}\tiny{$\pm 0.08$} \\
\midrule
\textit{Improvement} & \(+\mathbf{1.02}\) & \(+\mathbf{0.96}\) & \(+\mathbf{1.80}\) \\
\bottomrule
\end{tabular}

\end{minipage}
\hspace{0.1em}
\begin{minipage}[t]{0.48\textwidth}
\centering
\renewcommand{\arraystretch}{1.32}
\setlength{\tabcolsep}{0.1pt}
\footnotesize
\captionsetup{font = small}
\captionof{table}{Final accuracy (\%) for varying numbers of clients. Improvement offered by \texttt{FedRPCA} increases as the number of clients increases, i.e. number of updates to be merged increases.}
\label{tab:client-count}
\begin{tabular}{lccc}
\toprule
\textbf{Method} & \textbf{20} & \textbf{50} & \textbf{100} \\
\midrule
\texttt{FedAvg} & $92.81$\tiny{$\pm 0.07$} & $91.58$\tiny{$\pm 0.06$} & $88.62$\tiny{$\pm 0.09$} \\
\texttt{FedProx} & $92.78$\tiny{$\pm 0.07$} & $91.56$\tiny{$\pm 0.07$} & $88.61$\tiny{$\pm 0.09$} \\
\texttt{SCAFFOLD} & $90.72$\tiny{$\pm 0.07$} & $89.11$\tiny{$\pm 0.12$} & $87.95$\tiny{$\pm 0.02$} \\
\texttt{MOON} & $93.02$\tiny{$\pm 0.14$} & $91.57$\tiny{$\pm 0.06$} &  $88.77$\tiny{$\pm 0.08$} \\
\texttt{TIES-Merging} & $90.72$\tiny{$\pm 0.43$} & $89.51$\tiny{$\pm 1.10$} & $87.63$\tiny{$\pm 1.21$} \\
\texttt{Task Arithmetic} & $91.89$\tiny{$\pm 0.94$} & $91.48$\tiny{$\pm 0.58$} & $87.46$\tiny{$\pm 1.08$} \\
\rowcolor{lightgray}
\texttt{FedRPCA} & \bm{$93.11$}\tiny{$\pm 0.06$} & \bm{$92.40$}\tiny{$\pm 0.11$} & \bm{$90.30$}\tiny{$\pm 0.10$} \\
\midrule
\textit{Improvement} & \(+\mathbf{0.09}\) & \(+\mathbf{0.82}\) & \(+\mathbf{1.68}\) \\
\bottomrule
\end{tabular}
\end{minipage}

\paragraph{Impact of Rank.}
\begin{wraptable}{r}{0.6\linewidth}
    \centering
    \setlength{\tabcolsep}{1.5pt}
    \renewcommand{\arraystretch}{1.32}
    \footnotesize
    \captionsetup{font = small}
    \caption{Final accuracy (\%) and number of rounds required to reach $90\%$ accuracy (R@90) across rank $r = 8$ and rank $r = 32$. Bottom row shows \texttt{FedRPCA}'s improvement over second-best in accuracy and relative speed-up. While improvement in accuracy decreases as the rank increase, the relative speed-up remains consistent across different ranks. }
    \label{tab:rank-acc-rounds}
    \begin{tabular}{lcccc}
    \toprule
    \textbf{Method} & \multicolumn{2}{c}{\textbf{Rank 8}} & \multicolumn{2}{c}{\textbf{Rank 32}} \\
    \cmidrule(lr){2-3} \cmidrule(lr){4-5}
    & Acc. & R@90 & Acc. & R@90 \\
    \midrule
    \texttt{FedAvg} & $92.63$\tiny{$\pm 0.04$} & $41$  & $94.03$\tiny{$\pm 0.02$} & $19$ \\
    \texttt{FedProx} & $92.61$\tiny{$\pm 0.04$} & $41$  & $94.02$\tiny{$\pm 0.03$} & $19$ \\
    \texttt{SCAFFOLD} & $91.04$\tiny{$\pm 0.05$} & $86$  & $90.82$\tiny{$\pm 0.03$} & $107$ \\
    \texttt{MOON} & $92.72$\tiny{$\pm 0.02$}  & 38  & $93.96$\tiny{$\pm 0.03$} & 17 \\
    \texttt{TIES-Merging} & $90.41$\tiny{$\pm 1.25$} & $68$  & $91.54$\tiny{$\pm 1.29$} & $53$ \\
    \texttt{Task Arithmetic} & $92.47$\tiny{$\pm 0.46$} & $58$  & $93.71$\tiny{$\pm 1.43$} & $28$ \\
    \rowcolor{lightgray}
    \texttt{FedRPCA} & \bm{$93.30$}\tiny{$\pm 0.11$} & 25 & \bm{$94.05$}\tiny{$\pm 0.03$} & 12 \\
    \bottomrule
    \rowcolor{gray!10}
    \textit{Improvement} &
    \bm{$+0.58$} & \bm{$1.52\times$} &
    \bm{$+0.02$} & \bm{$1.42\times$} \\
    \bottomrule
    \end{tabular}
\end{wraptable}
\Cref{tab:rank-acc-rounds} shows the impact of changing rank on the performance of \texttt{FedRPCA} and other baselines. As the rank increases, the model's capacity also increases, bringing it closer to the expressiveness of full fine-tuning. In this regime, prior work has shown that the adverse effects of data heterogeneity on \texttt{FedAvg} are reduced \cite{NguyenWMSR23, Chen0LSC23}. Consequently, we observe that the final accuracy gap between \texttt{FedRPCA} and \texttt{FedAvg} narrows at higher ranks. Nonetheless, \texttt{FedRPCA} continues to offer a consistent speed-up in convergence. This is reflected in the R@90 column of \Cref{tab:rank-acc-rounds}, which reports the number of rounds required to reach $90\%$ of the final accuracy – which remains stable across ranks and demonstrates the efficiency of our method even when the performance gap closes.
\section{Conclusion}
\label{sec:conclusion}
In this paper we propose \texttt{FedRPCA}, a server-side aggregation algorithm for federated parameter-efficient fine-tuning of pretrained models using \texttt{LoRA}. \texttt{FedRPCA} builds on a key insight that client updates have a common signal and a client-specific signal -- it uses \texttt{Robust-PCA} to separate these signals, and applies different scaling factors to them when averaging the updates at the server. It accelerates convergence and boosts accuracy, without modifying the client-side objective or optimization procedure. Future directions include theoretical analysis of the proposed method, and application to other parameter-efficient fine-tuning techniques beyond \texttt{LoRA}.

\section{Acknowledgements}
This work was supported in part by NSF grants CCF 2045694, CNS-2112471, CPS-2111751, SHF-2107024, ONR grant N00014-23-1-2149, a Google Research Scholar Award, the CyLab IoT Enterprise Security Initiative, the David and La Verne Barakat fellowship and the Benjamin Garver Lamme/Westinghouse Fellowship. This work used Bridges-2 GPU at the Pittsburgh Supercomputing Center through allocation CIS240784 from the Advanced Cyberinfrastructure Coordination Ecosystem: Services \& Support (ACCESS) program, which is supported by NSF grants \#2138259, \#2138286, \#2138307, \#2137603, and \#2138296 \cite{access}.

\bibliographystyle{plain}
\bibliography{ref}

\appendix

\newpage

\begin{center}
    {\LARGE \textbf{Appendix}}
\end{center}

\vspace{1em}

\section*{Organization of Appendix}
\startcontents[sections]
\printcontents[sections]{l}{1}{\setcounter{tocdepth}{2}}

\clearpage

\section{Additional Related Work.}
\label{appendix:related_work}

\paragraph{Federated Fine-Tuning.} Several recent works have examined the role of pre-trained initialization in FL, including theoretical explanations of its generalization benefits \cite{jhunjhunwala2025initialization}, the impact of wireless bandwidth constraints \cite{wang2025federated}, and deployment on extreme edge hardware \cite{woisetschlager2024federated}. \cite{chen2022fedtune, peng2024fedpft, chen2024feddat, wang2024one} show that adapter-based or low-rank updates enable effective fine-tuning with significantly fewer communication rounds. Several studies also modify the \texttt{LoRA} mechanism itself to suit FL. \texttt{SLoRA} \cite{babakniya2023slora} proposes to initialized the $\bA$ and $\bB$ matrices as a sparse approximation of the true full-model updated $\Delta \bW$. \texttt{pFedLoRA} \cite{yi2023pfedlora} tackles model heterogeneity by giving each client its own LoRA rank and then learning a meta-network that generates personalized adapter weights. \texttt{FDLoRA} \cite{qi2024fdlora} trains both global and personalized \texttt{LoRA} modules at each client, where only the parameters of the global module are communicated and averaged across clients. \texttt{SA-FedLoRA} \cite{yang2024sa} proposed to adaptively change the rank of the \texttt{LoRA} parameters during training motivated by Simulated Annealing, and \texttt{FedALT} \cite{bian2025fedalt} emphasize local training of \texttt{LoRA} parameter by letting clients continue training on their local models from the previous round and instead use “rest-of-the-world” regularizer that discourages drift from the global representation. \texttt{FFALoRA} \cite{FFALoRA} proposed to freeze the $\bA$ parameters and only train the $\bB$ parameters to achieve exact aggregation \cite{singhal2025fed} and also improve privacy guarantees. \texttt{DP-LoRA} \cite{liu2025differentially} provides a formal privacy guarantee by introducing random Gaussian noise in the low-rank updates before aggregation.

\paragraph{Merging \texttt{LoRA} Adapters.}

Recent research on composing \texttt{LoRA} adapters in the centralized setting can be broadly categorized into three threads. Weight-space fusion methods merge independently fine-tuned adapters through averaging, clustering, or simple arithmetic. \texttt{AdapterSoup} \cite{chronopoulou-etal-2023-adaptersoup} trains a set of domain-specific adapters and selects which ones to average at test time for each novel domain. \texttt{LoRA Soups} \cite{LoRASoup} demonstrate that concatenating LoRAs using optimally learned weights outperforms traditional model and data-merging techniques. \texttt{LoRA-Lego} \cite{LegoLoraMerge} introduces the idea of Minimal Semantic Units (MSUs), where each LoRA rank acts as an independent unit. These MSUs exhibit permutation invariance and concatenation-summation equivalence, enabling flexible composition strategies. A second line of work uses optimization-based or learned merging techniques. \texttt{ZipLoRA} \cite{shah2023ZipLoRA} formulates adapter fusion as an optimization problem, identifying disjoint merger coefficients to blend LoRAs while preserving both subject and style. \texttt{LoRA-rar} \cite{shenaj2024lora} employs a hypernetwork to generate a single adapter conditioned on both, allowing for end-to-end learned merging. \texttt{PETA} \cite{PETA} uses partial linearization when doing \texttt{LoRA} optimization to enable efficient merging. \texttt{KnOTS} \cite{knots} and \texttt{LoRM} \cite{salamiclosed} propose merging schemes grounded in the optimization dynamics of LoRA training. Zhang et al. \cite{zhang2023composing} interpret adapters as vectors in weight space and compose new skills using scaled vector arithmetic. A third direction explores dynamic or mixture-of-experts (MoE) approaches, which retain multiple adapters and learn to select or weight them at inference time. Methods such as \texttt{Mixture-of-LoRAs} \cite{LoRAMixture}, \texttt{X-LoRA} \cite{X-LoRA}, \texttt{MOLE} \cite{MOLE}, \texttt{LoRA-Flow} \cite{wang-etal-2024-lora-flow}, and \texttt{S-LoRA} \cite{sheng2023slora} achieve strong performance but often incur additional memory and routing overhead.

\newpage

\section{Additional Algorithmic Details}

\subsection{Implementing \texttt{Robust-PCA}}
\label{appendix:robust_pca_implementation}

To implement \texttt{Robust-PCA}, we adopt the Alternating Direction Method of Multipliers (ADMM) framework \cite{boyd2011distributed}, following the algorithm proposed in \cite{candes2011robust} (see Algorithm 1 in \cite{candes2011robust}). Given an input matrix $\bM \in \mathbb{R}^{d_1 \times d_2}$, we use the default hyperparameter values commonly recommended in the literature: $\lambda = 1 / \sqrt{\max(d_1, d_2)}$, which controls the sparsity of the matrix $\bS$, and $\mu = d_1 \cdot d_2 / (4||\bM||_1)$, which can be interpreted as an effective step size or learning rate. These choices have been found to be robust across a range of problem settings and work well out of the box in our experiments as well. \Cref{alg:robust_pca} provides a PyTorch-style pseudocode implementation of the \texttt{Robust-PCA} algorithm used in our setup.

\lstdefinestyle{pytorch}{
  language=Python,
  basicstyle=\ttfamily\small,
  keywordstyle=\color{blue}\bfseries,
  commentstyle=\color{gray},
  showstringspaces=false,
  tabsize=4,
  numbers=none,
  numberstyle=\tiny,
  frame=none,
  framerule=0.2pt,
  rulecolor=\color{black!30},
  breaklines=true
}

\begin{algorithm}[H]
\caption{\texttt{Robust-PCA} via ADMM (PyTorch-like)}
\label{alg:robust_pca}
\begin{lstlisting}[style=pytorch]
def robust_pca(M: Tensor,
               mu: float | None = None,
               lam: float | None = None,
               tol: float = 1e-7,
               max_iter: int = 1000) -> tuple[Tensor, Tensor]:
    """Decompose D into low-rank L and sparse S (M = L + S)."""
    mu  = mu  or  M.numel() / (4 * M.abs().sum())
    lam = lam or 1.0 / torch.sqrt(torch.tensor(max(M.shape)))
    rho = 1.0 / mu

    S, Y = torch.zeros_like(M), torch.zeros_like(M)

    shrink = lambda X, t: torch.sign(X) * torch.clamp(X.abs() - t, min=0.0)

    def svt(X, t):
        U, s, Vh = torch.linalg.svd(X, full_matrices=False)
        s = shrink(s, t)
        return (U * s) @ Vh

    for _ in range(max_iter):
        L = svt(M - S + rho * Y,  rho)          # low-rank update
        S = shrink(M - L + rho * Y, rho * lam)  # sparse update
        Y = Y + mu * (M - L - S)                # dual update
        if (M - L - S).norm() <= tol * M.norm():
            break
    return L, S
\end{lstlisting}
\end{algorithm}

\subsection{Computational Efficiency of \texttt{FedRPCA}}
\label{appendix:compute_efficiency}

As outlined in \Cref{algo1}, we apply the \texttt{Robust-PCA} subroutine individually to the $\bA$ and $\bB$ \texttt{LoRA} parameters. We also apply this jointly across the $(\bA, \bB)$ pairs in each layer of the model. As shown in \Cref{alg:robust_pca}, this subroutine involves repeated SVD operations. Nonetheless, since the input matrices $\bM$ to the \texttt{Robust-PCA} algorithm are typically of size $\sim 10^3$, we find that the corresponding computation is relatively lightweight at the server. In practice, the majority of compute overhead stems from the local optimization steps used to update the $\bA$ and $\bB$ matrices.

\begin{wrapfigure}{r}{0.35\textwidth}
\centering 
\includegraphics[width=0.35\textwidth]{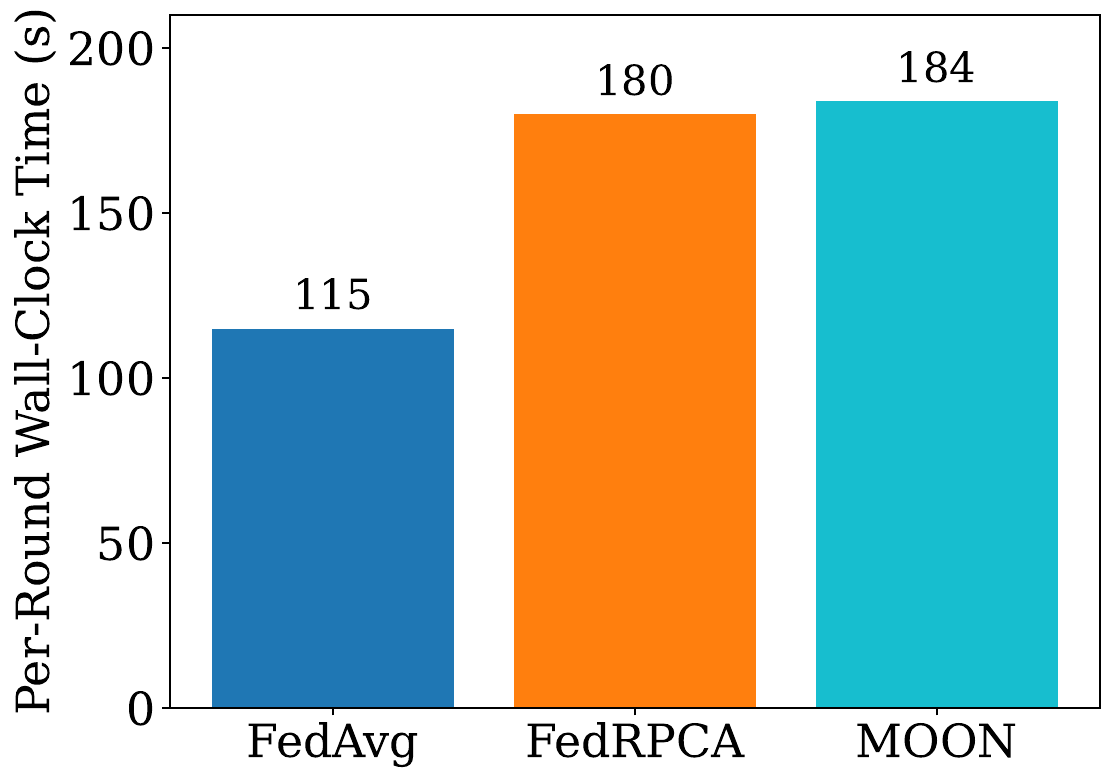}
\captionsetup{font = small} 
\vspace{-2.0em}
\caption{Per-round wall-clock times for different algorithms when fine-tuning ViT-B/32 on SVHN dataset distributed across 50 clients.}
\vspace{-6em}
\label{fig:wallclock}
\end{wrapfigure}
\Cref{fig:wallclock} compares the per-round wall-clock time of \texttt{FedAvg}, \texttt{FedRPCA}, and \texttt{MOON} for the our FL on fine-tuning ViT-B/32 on the SVHN dataset distributed across 50 clients. We observe that \texttt{FedRPCA} incurs only a modest increase in compute time – approximately $1.5\times$ over \texttt{FedAvg}, and remains comparable to algorithms such as \texttt{MOON}, which instead modifies the local training procedure.

Future work can further reduce this overhead by parallelizing \texttt{Robust-PCA} computations across layers and modules, as well as by leveraging more efficient \texttt{Robust-PCA} implementations (e.g., \cite{OnlineRPCA, GoDec, GoDec+}).

\subsection{Adaptive scaling in \texttt{FedRPCA}}
\label{appendix:adaptive_scaling}

\begin{figure}[h]
  \centering
  
  \begin{subfigure}[b]{0.31\linewidth}
    \includegraphics[width=\linewidth]{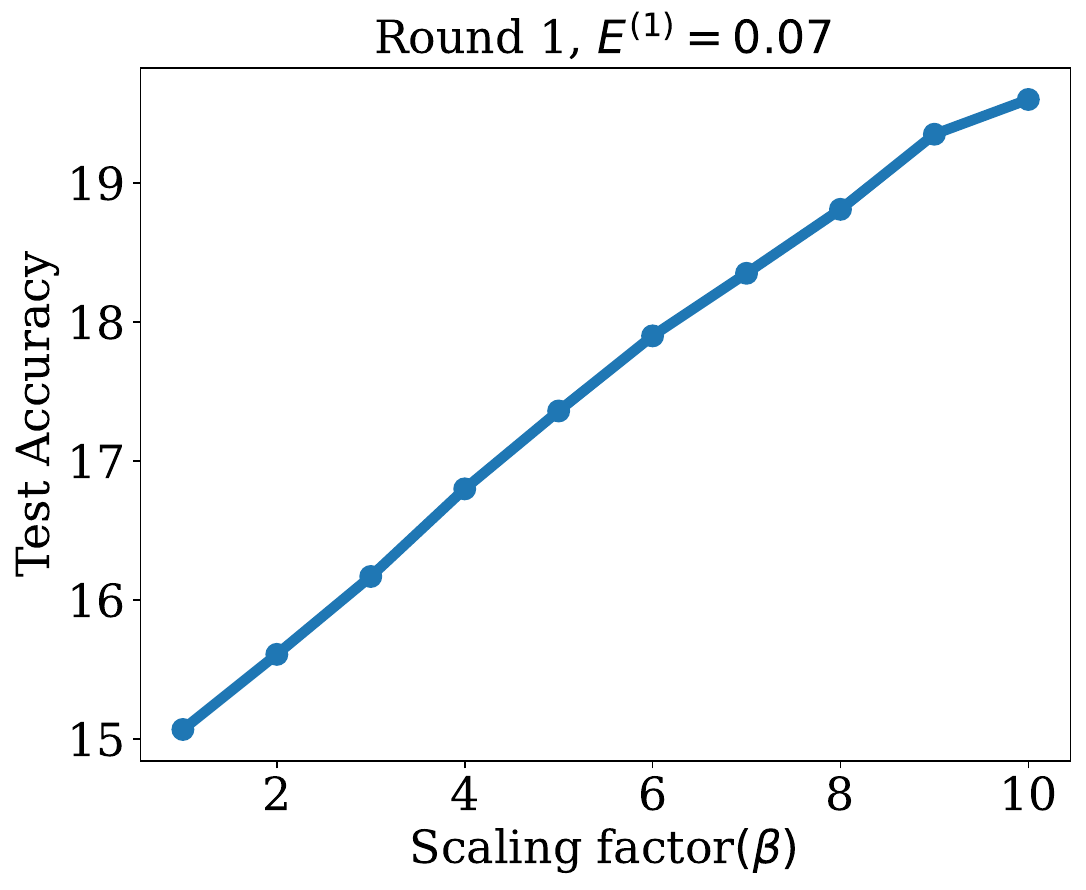}
    \caption{}
  \end{subfigure}
  \begin{subfigure}[b]{0.33\linewidth}
    \includegraphics[width=\linewidth]{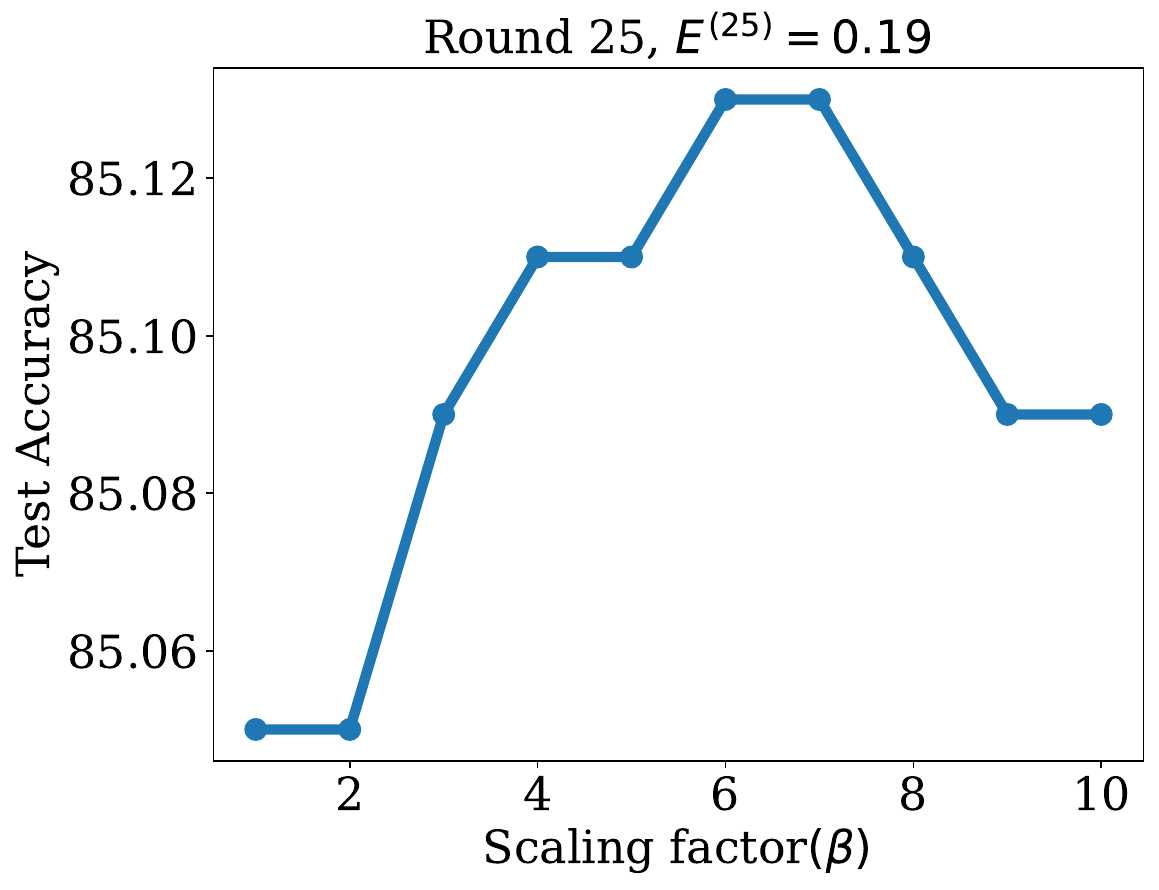}
    \caption{}
  \end{subfigure}
  \begin{subfigure}[b]{0.33\linewidth}
    \includegraphics[width=\linewidth]{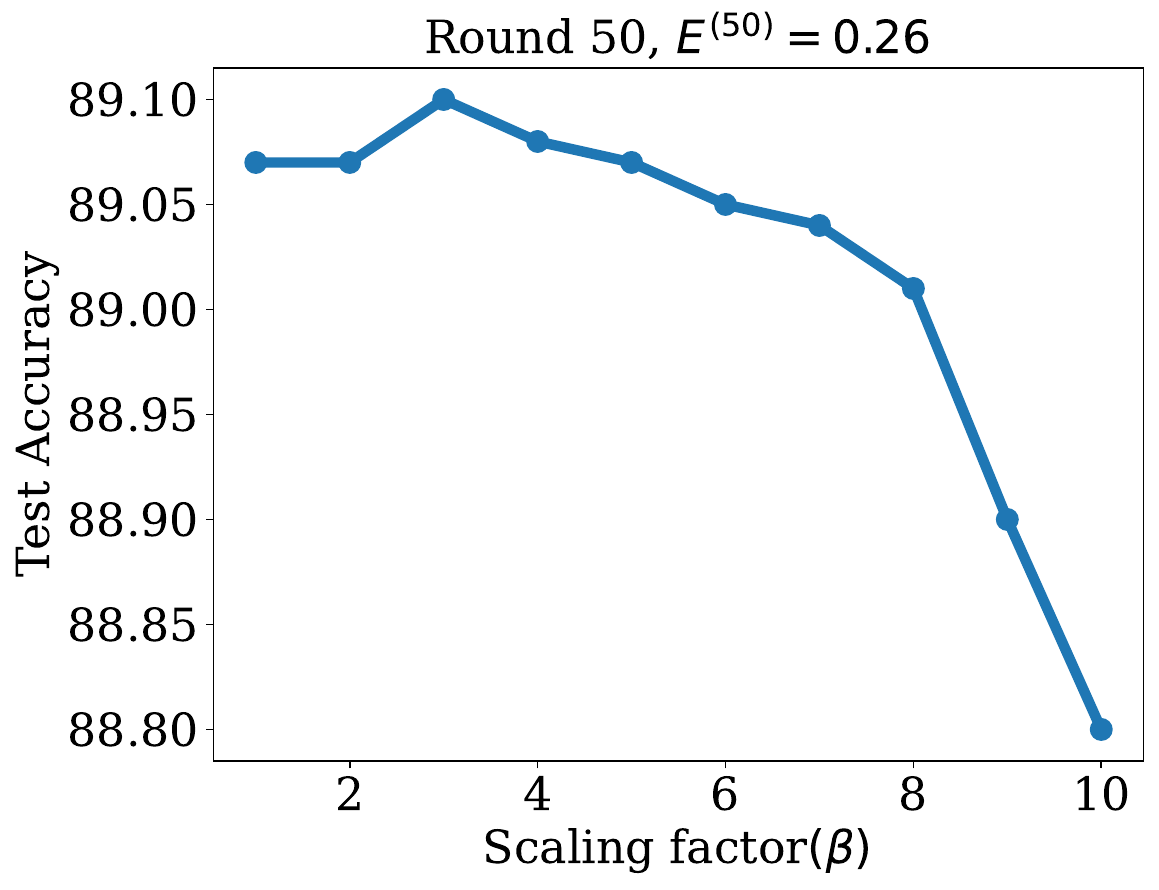}
    \caption{}
  \end{subfigure}

  \caption{Performance of \texttt{FedRPCA}-style aggregation for different values of the scaling factor $\beta$ at various training rounds when fine-tuning a ViT-B/32 model on the SVHN dataset with 50 clients. At each round $t$, $E^{(t)}$ denotes the average of $E_{\bA}^{(t)}$ and $E_{\bB}^{(t)}$, computed across all layers. We observe that the optimal choice of $\beta$ generally decreases over training and is approximately inversely proportional to $E^{(t)}$.}
  \label{fig:opt_beta}
\end{figure}

\begin{wrapfigure}{r}{0.4\textwidth}
\centering 
\vspace{-1.8em}
\includegraphics[width=0.4\textwidth]{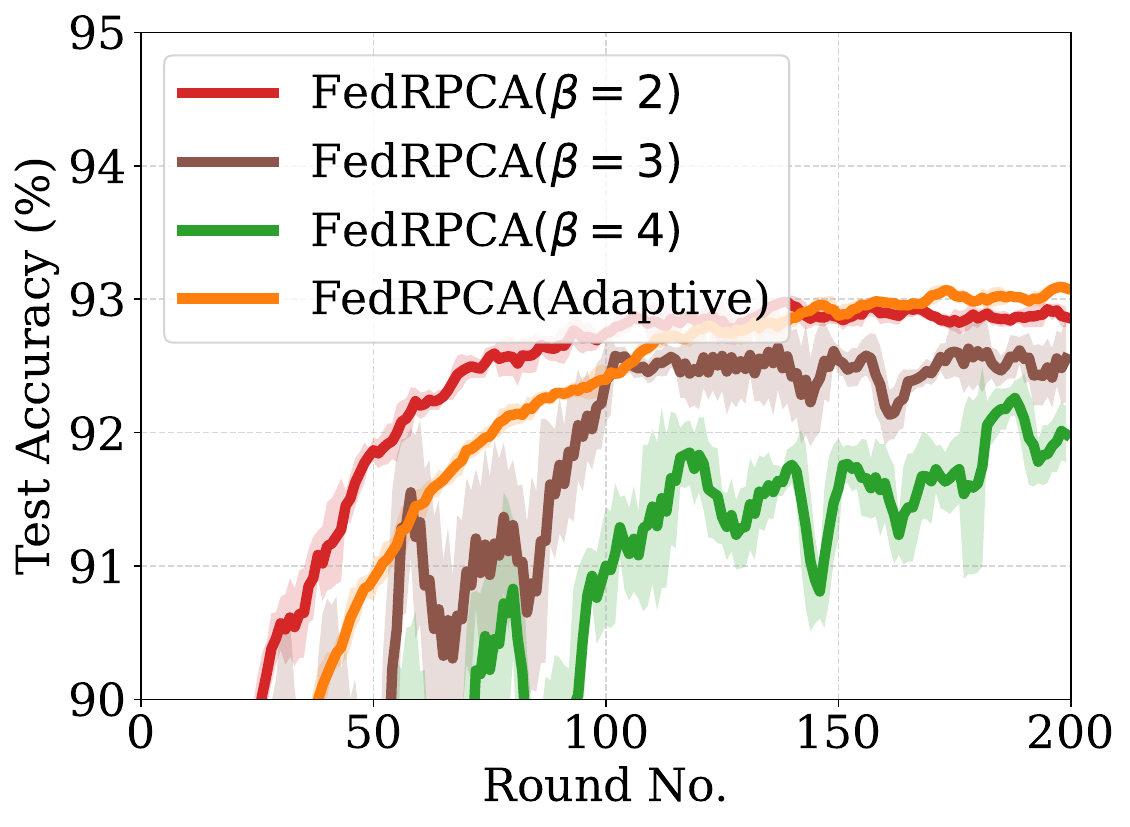}
\captionsetup{font = small} 
\caption{Performance of \texttt{FedRPCA} with different fixed $\beta$ values and the proposed adaptive $\beta$ schedule when fine-tuning ViT-B/32 on SVHN dataset distributed across 50 clients. The proposed adaptive schedule achieves higher final accuracy than all fixed choice of $\beta$.}
\label{fig:rpca_adapt_beta}
\vspace{-7em}
\end{wrapfigure}
As discussed in \Cref{sec:method}, we find that the optimal value of $\beta$ for \texttt{FedRPCA} varies throughout training. We demonstrate this in \Cref{fig:opt_beta} by running \texttt{FedAvg} on the SVHN dataset for 100 rounds, and measuring the optimal choice of $\beta$ assuming we were instead performing \texttt{FedRPCA}-style aggregation at rounds 1, 25, and 50. We observe that the optimal $\beta$ generally decreases as training progresses and appears roughly inversely proportional to the norm contribution of the sparse updates (obtained via \texttt{Robust-PCA}) to the original updates, as measured by 
$E_{\bA}^{(t)} = \frac{\| \bS_{\bA}^{(t)} \cdot \bm{1} \|}{\| \bM_{\bA}^{(t)} \cdot \bm{1} \|},$ (with a similar definition for $\bB$). Motivated by this, we propose a simple heuristic: set $\beta_{\bA}^{(t)} = 1 / E_{\bA}^{(t)}$ and $\beta_{\bB}^{(t)} = 1 / E_{\bB}^{(t)}$ in every round. As shown in \Cref{fig:rpca_adapt_beta} this adaptive schedule outperforms fixed choices of $\beta$ for $\beta \in \{2, 3, 4\}$. Future work can explore further improvements to this heuristic schedule.

\newpage

\section{Additional Experimental Details and Results}
\label{appendix:additional_experiments}
\subsection{Hyperparameters}
\label{appendix:hyperparameters}
In this section, we describe the hyperparameters used in our experiments. Table \ref{tab:training_params} highlights the fixed hyperparameters for our FL fine-tuning procedure. Recall that we use a \textbf{CLIP ViT-B/32} for all our vision tasks, \textbf{GPT-2} for 20News and \textbf{T5-Base} for MRQA. 

\begin{table}[ht]
\centering
\caption{Training configuration for each model.}
\begin{tabular}{l c c c}
\toprule
\multicolumn{1}{c}{} & \multicolumn{3}{c}{\textbf{Model}} \\
\cmidrule(lr){2-4}
 & \textbf{CLIP ViT‑B‑32} & \textbf{GPT‑2} & \textbf{T5‑Base} \\
\midrule
Batch Size            & $32$ & $32$ & $32$ \\
Local Epochs          & $1$ & $1$ & $1$ \\
Max Sequence Length   & ‒‒ & $256$ & $256$ \\
Optimizer             & \textsc{AdamW} & \textsc{AdamW} & \textsc{AdamW} \\
Learning Rate         & $0.0001$ & $0.001$ & $0.0001$ \\
Weight Decay          & $0.1$ & $0.1$ & $0.1$ \\
\bottomrule
\end{tabular}
\label{tab:training_params}
\end{table}

For the following baselines, we perform a grid search of the method's primary tuning parameter. In particular, we test the following: 
\begin{itemize}
     \item \texttt{TIES-Merging}: Grid search over the sparsity ratio $s$. We test values $s=\{0.1, 0.3, 0.5, 0.7, 0.9\}$
    \item \texttt{FedProx}: Grid search over the proximal coefficient $\mu$. We test values $\mu=\{0.001, 0.01,\\  0.1, 1.0\}$
    \item \texttt{MOON}: Grid search over the contrastive regularization weight $\mu$. We test values $\mu=\{0.001, 0.01,0.1, 1.0\}$
    \item \texttt{Task Arithmetic}: Grid search over the scaling factor applied to the task vectors $\beta$. We test values $\beta=\{2,3,4\}$
    
\end{itemize}
Tables \ref{tab:vision_opt_hparams} and \ref{tab:language_opt_hparams} describe the optimal hyperparameter choices for each benchmark in the vision datasets and language datasets, respectively. 
\begin{table}[ht]
\centering
\caption{Optimal hyper‑parameter values for vision benchmarks.}
\resizebox{\textwidth}{!}{%
\begin{tabular}{l c c c c}
\toprule
\multicolumn{1}{c}{} & \multicolumn{4}{c}{\textbf{Model / Dataset}} \\
\cmidrule(lr){2-5}
\textbf{Method} 
& \textbf{CLIP ViT‑B‑32 / SVHN} 
& \textbf{CLIP ViT‑B‑32 / EuroSAT} 
& \textbf{CLIP ViT‑B‑32 / DTD} 
& \textbf{CLIP ViT‑B‑32 / Stanford Cars} \\
\midrule
\texttt{TIES-Merging} ($s$)        & $0.1$   & $0.9$   & $0.1$   & $0.9$   \\
\texttt{MOON} ($\mu$)           & $0.001$ & $1.0$   & $0.01$  & $0.001$ \\
\texttt{FedProx} ($\mu$)        & $0.01$  & $0.001$ & $0.01 $ & $0.01$  \\
\texttt{Task Arithmetic} ($\beta$) & $2$ & $2 $& $2  $   & $2 $    \\
\bottomrule
\end{tabular}}
\label{tab:vision_opt_hparams}
\end{table}
\begin{table}[ht]
\centering
\caption{Optimal hyper‑parameter values for language benchmarks.}
\begin{tabular}{l c c}
\toprule
\multicolumn{1}{c}{} & \multicolumn{2}{c}{\textbf{Model / Dataset}} \\
\cmidrule(lr){2-3}
\textbf{Method} 
& \textbf{GPT‑2 / 20News} 
& \textbf{T5‑Base / MRQA} \\
\midrule
\texttt{TIES-Merging} ($s$)        & $0.1 $ & $0.1$  \\
\texttt{MOON} ($\mu$)           & $0.01$ & $0.01$ \\
\texttt{FedProx} ($\mu$)        & $0.01$ & $0.01$ \\
\texttt{Task Arithmetic} ($\beta$) & $2  $  & $2$    \\
\bottomrule
\end{tabular}
\label{tab:language_opt_hparams}
\end{table}
\newpage

\subsection{Accuracy vs. Rounds Plots}
\label{appendix:acc_vs_round_plots}
Here we provide the test accuracy vs. number of rounds plots for the results presented in \Cref{sec:expts}.

\subsubsection{Main Results}

Below we present the test accuracy curves corresponding to \Cref{tab:accuracy-only}.

\begin{figure}[h]
  \centering
  {
  \includegraphics[height=0.34cm]{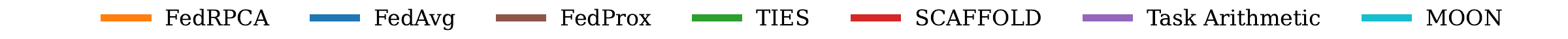}}

  \vspace{0.5em}    
  \begin{subfigure}[b]{0.32\linewidth}
    \includegraphics[width=\linewidth]{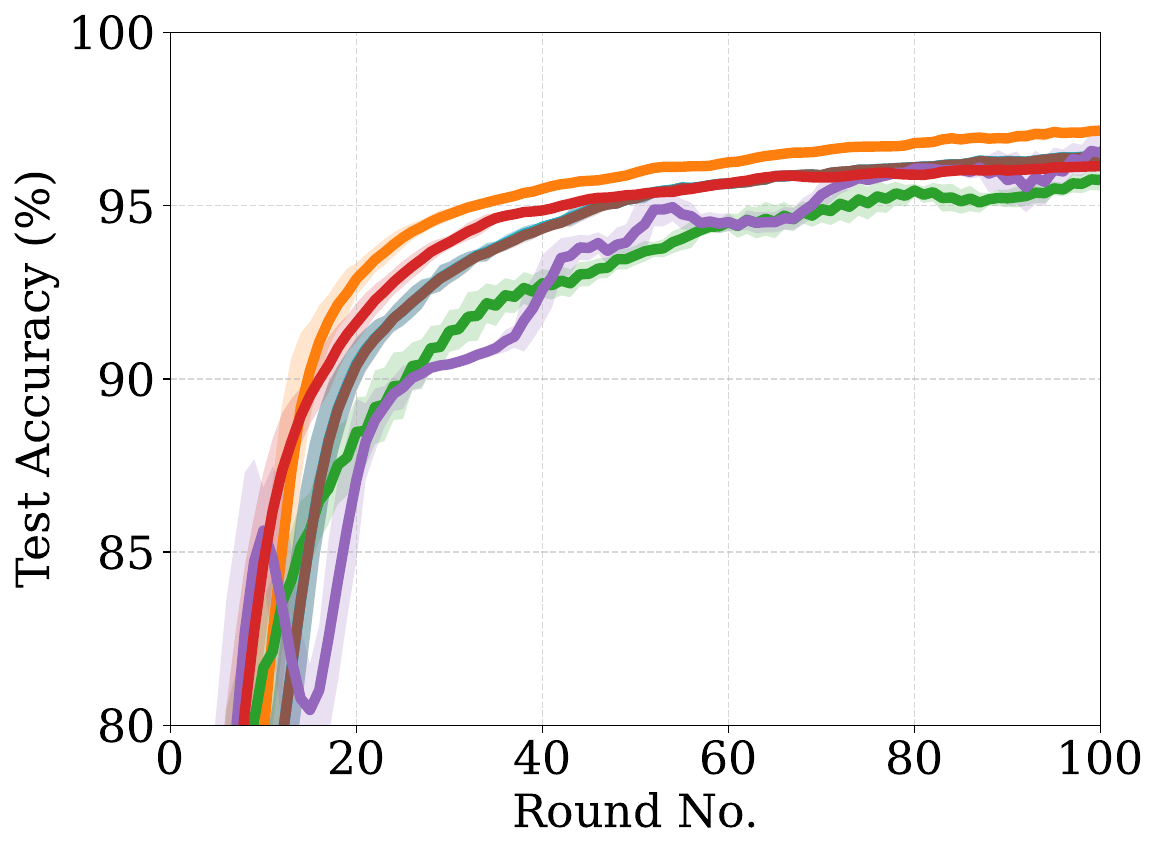}
    \caption{EuroSAT}
  \end{subfigure}
  \begin{subfigure}[b]{0.32\linewidth}
    \includegraphics[width=\linewidth]{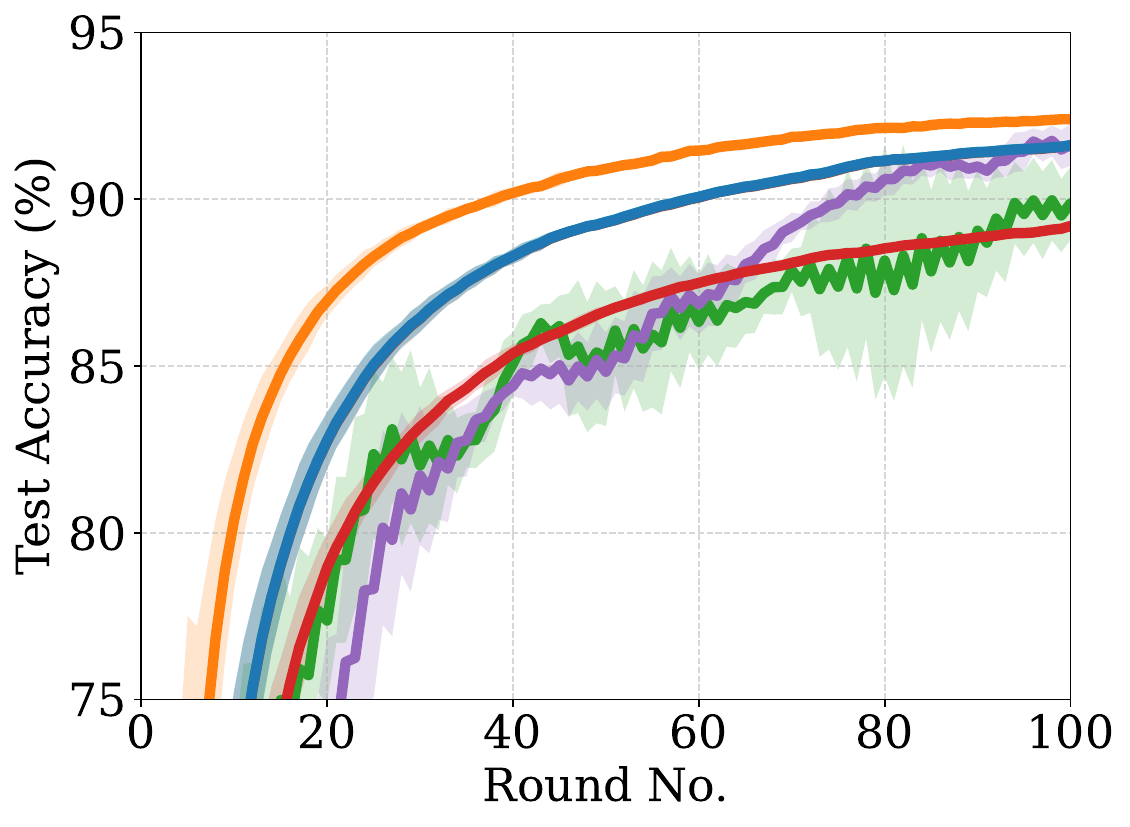}
    \caption{SVHN}
  \end{subfigure}
  \begin{subfigure}[b]{0.32\linewidth}
    \includegraphics[width=\linewidth]{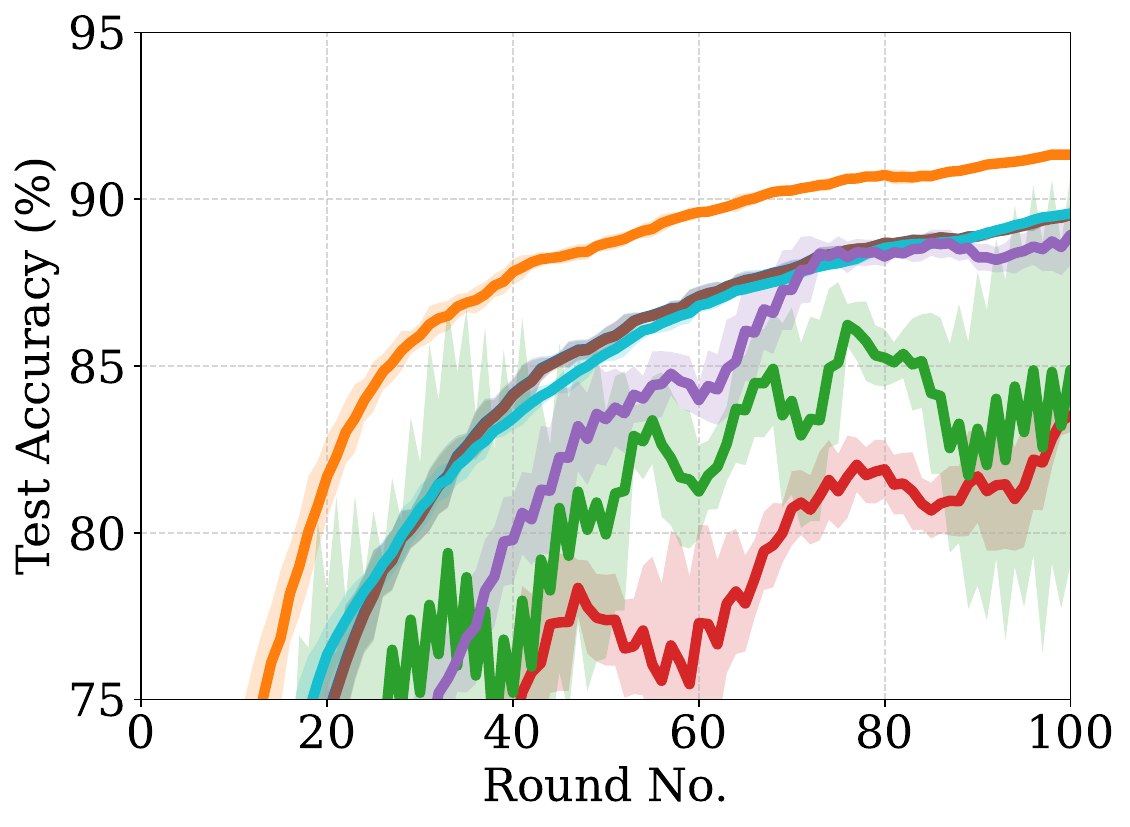}
    \caption{DTD}
  \end{subfigure}

  \begin{subfigure}[b]{0.32\linewidth}
    \includegraphics[width=\linewidth]{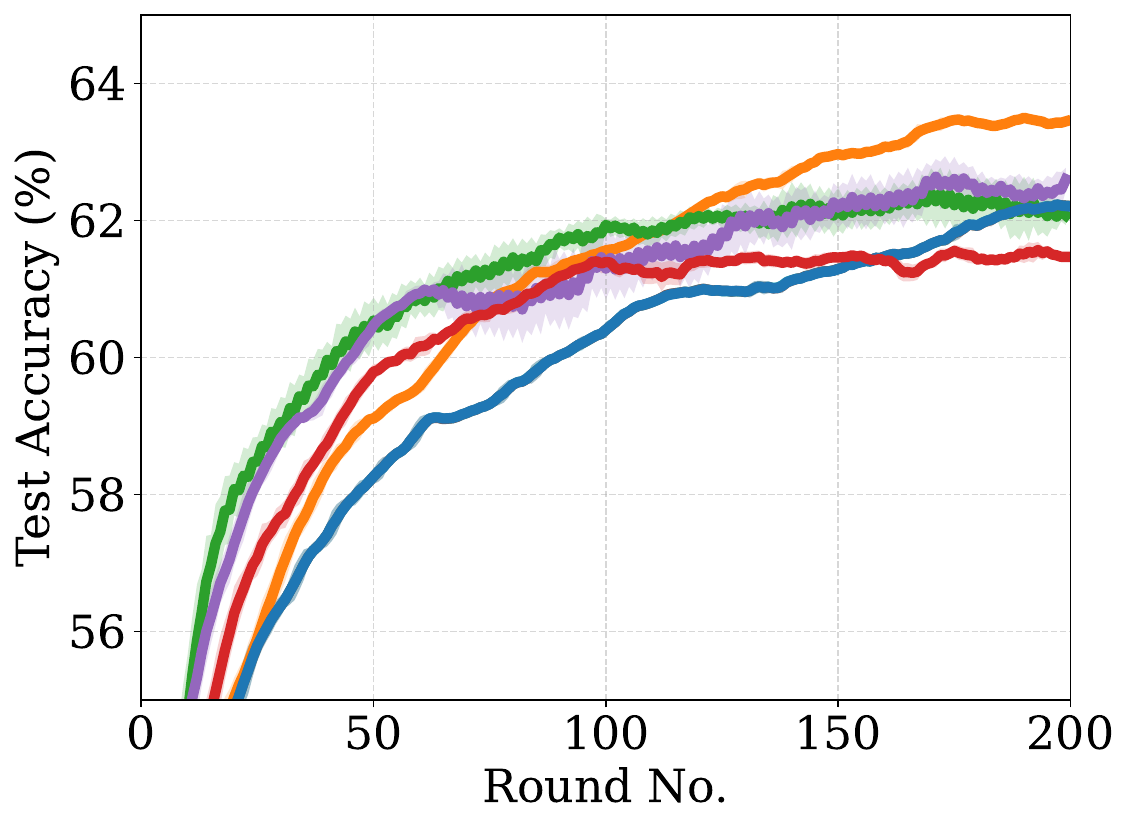}
    \caption{Stanford Cars}
  \end{subfigure}
  \begin{subfigure}[b]{0.32\linewidth}
    \includegraphics[width=\linewidth]{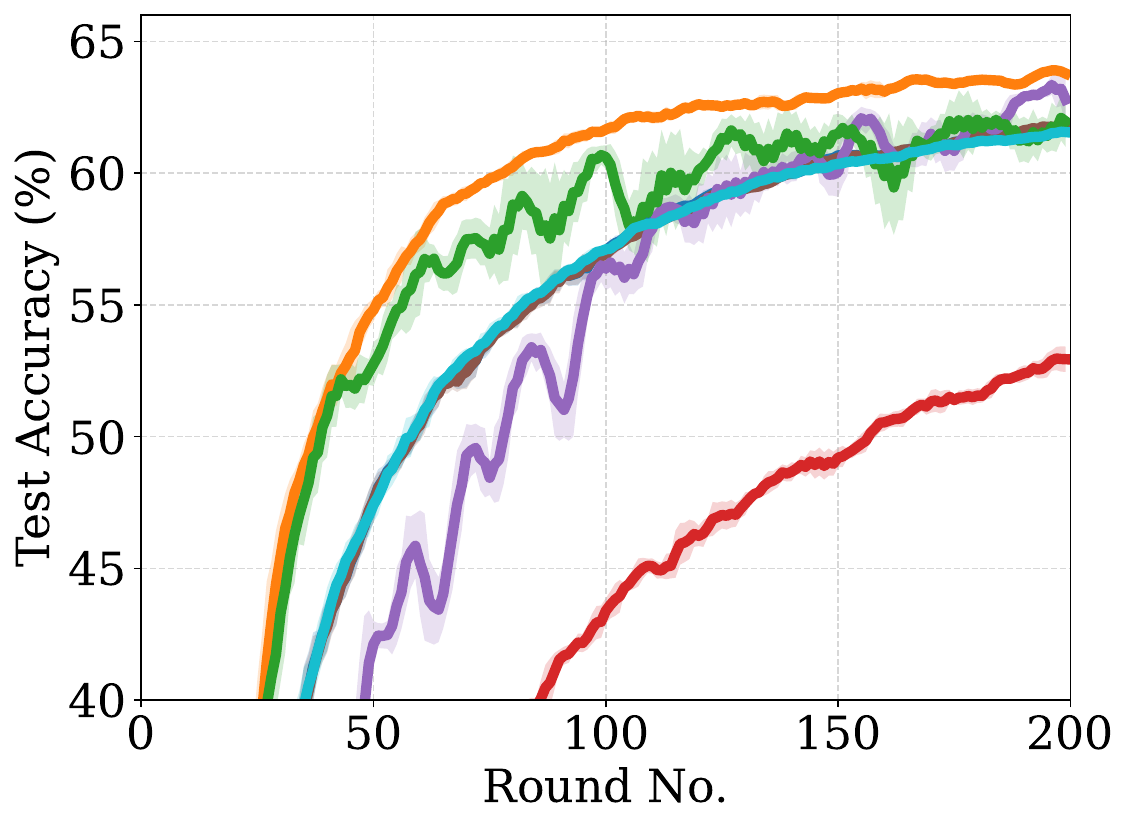}
    \caption{20News}
  \end{subfigure}
  \begin{subfigure}[b]{0.32\linewidth}
    \includegraphics[width=\linewidth]{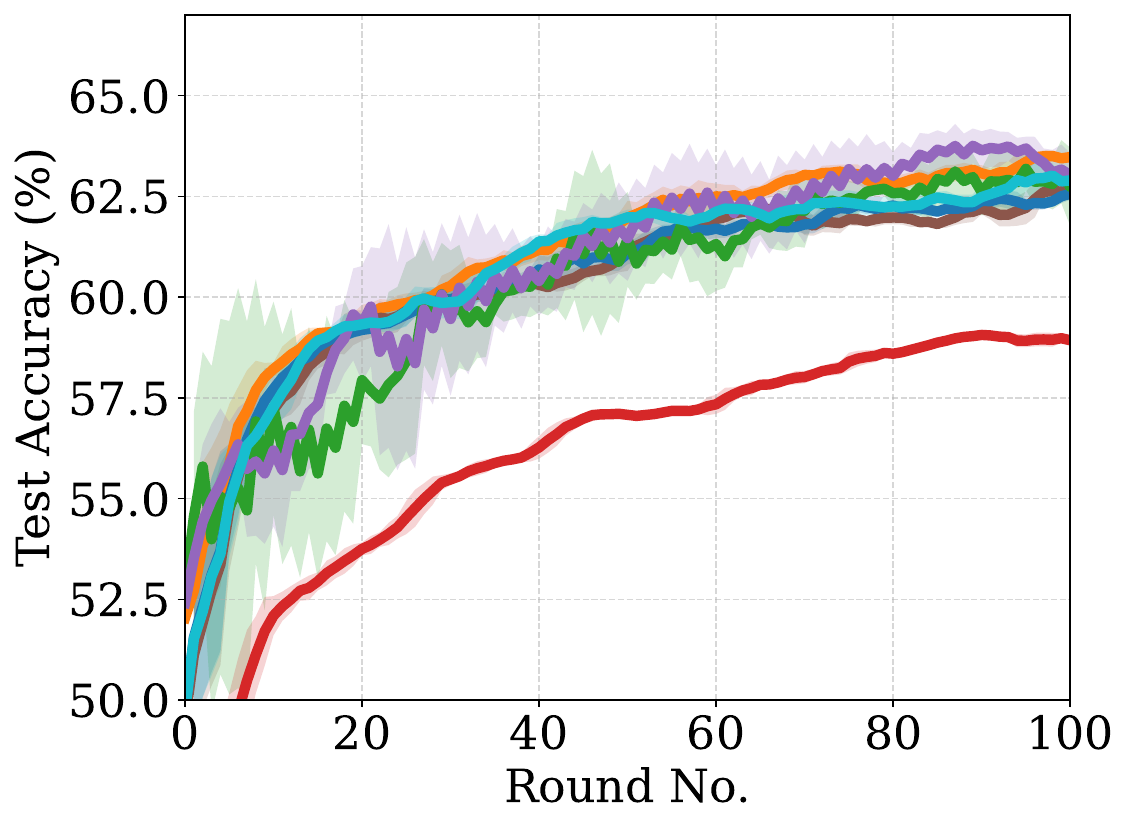}
    \caption{MRQA}
  \end{subfigure}

  \caption{Performance of \texttt{FedRPCA} and baselines across different vision and language datasets. \texttt{FedRPCA} consistently outperforms baselines, especially for harder datasets like DTD and 20News.}
  \label{fig:hetero_plots}
\end{figure}


\subsubsection{Ablation Results}
Below we present the test accuracy curves corresponding to \Cref{tab:heterogeneity} (\Cref{fig:hetero_plots}), \Cref{tab:client-count} (\Cref{fig:client_plots}) and \Cref{tab:rank-acc-rounds} (\Cref{fig:rank_plots}).

\begin{figure}[!h]
  \centering
  {
  \includegraphics[height=0.34cm]{figures/common_legend.pdf}} 
  \begin{subfigure}[b]{0.32\linewidth}
    \includegraphics[width=\linewidth]{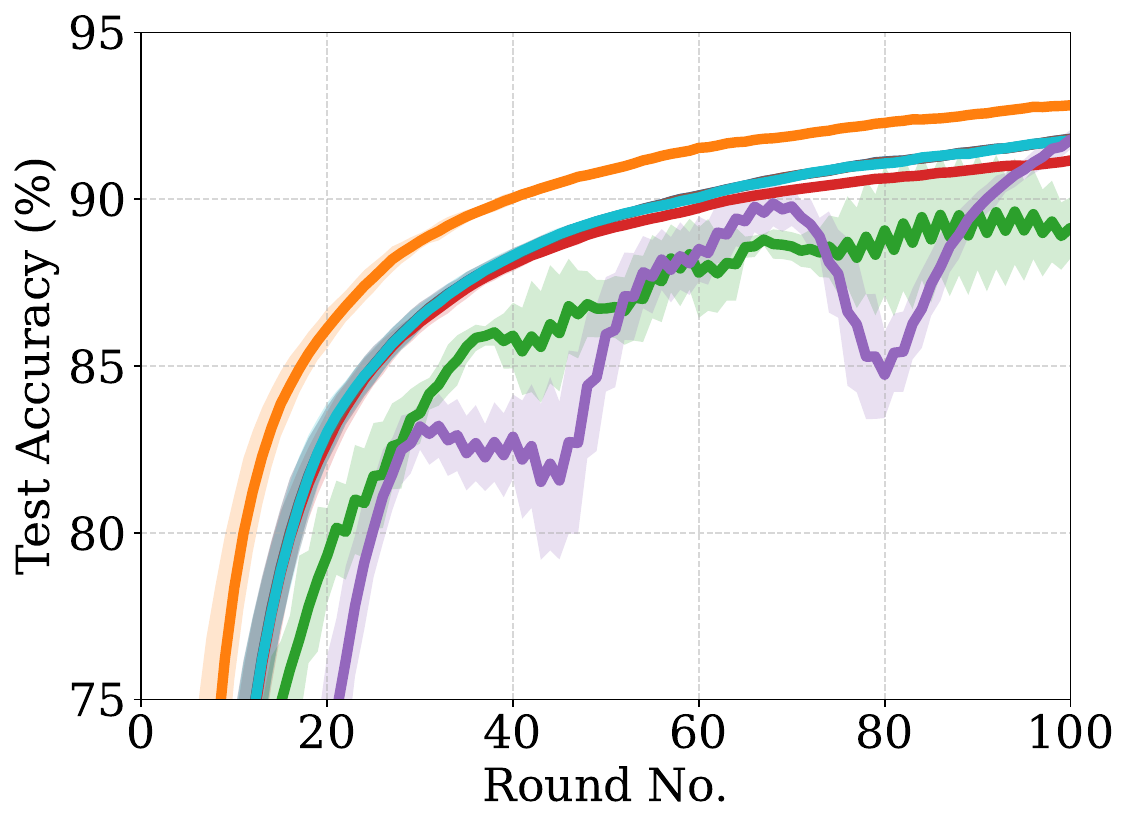}
    \caption{$\alpha = 10$}
  \end{subfigure}
  \begin{subfigure}[b]{0.32\linewidth}
    \includegraphics[width=\linewidth]{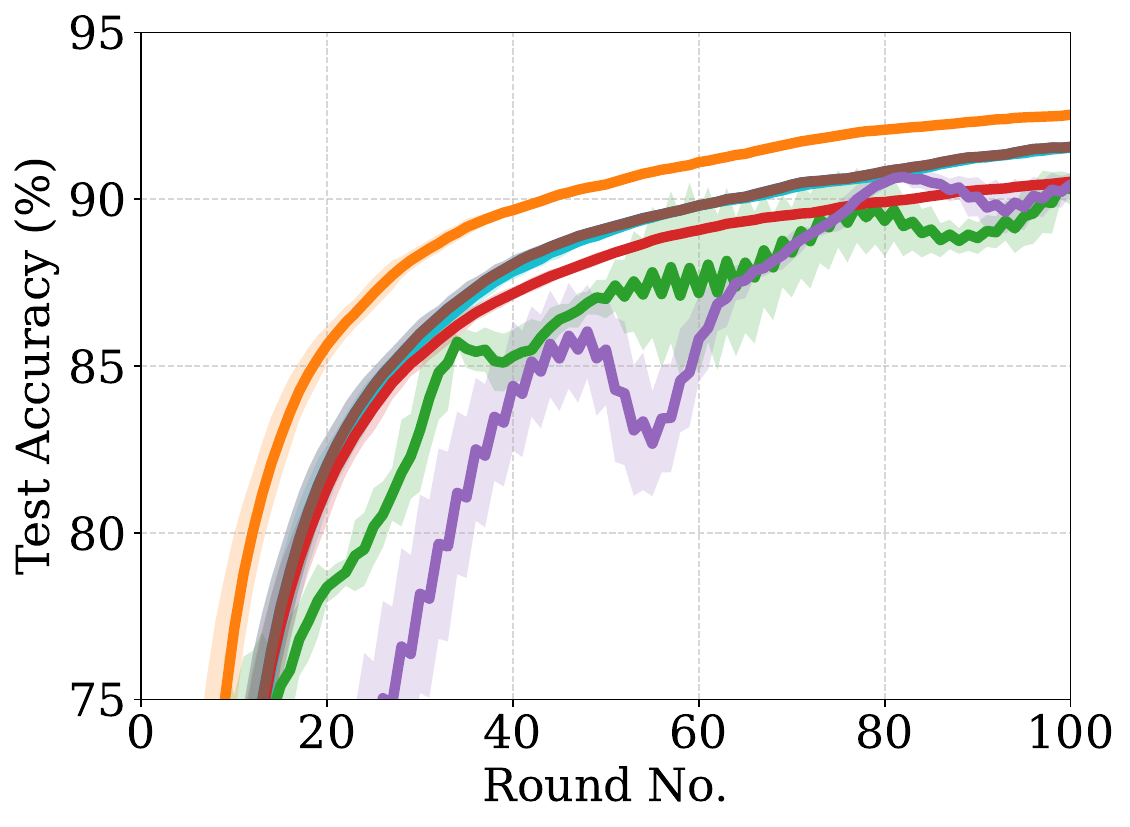}
    \caption{$\alpha = 1.0$}
  \end{subfigure}
  \begin{subfigure}[b]{0.32\linewidth}
    \includegraphics[width=\linewidth]{figures/svhn_alpha_0.1_acc_plot.pdf}
    \caption{$\alpha = 0.1$}
  \end{subfigure}
  \caption{Effect of heterogeneity ($\alpha$) on the performance of \texttt{FedRPCA} and baselines. Improvement offered by \texttt{FedRPCA} increases as heterogeneity increases.}
  \label{fig:hetero_plots}
\end{figure}
\begin{figure}[!h]
  \centering
  {
  \includegraphics[height=0.34cm]{figures/common_legend.pdf}}  
  \begin{subfigure}[b]{0.32\linewidth}
    \includegraphics[width=\linewidth]{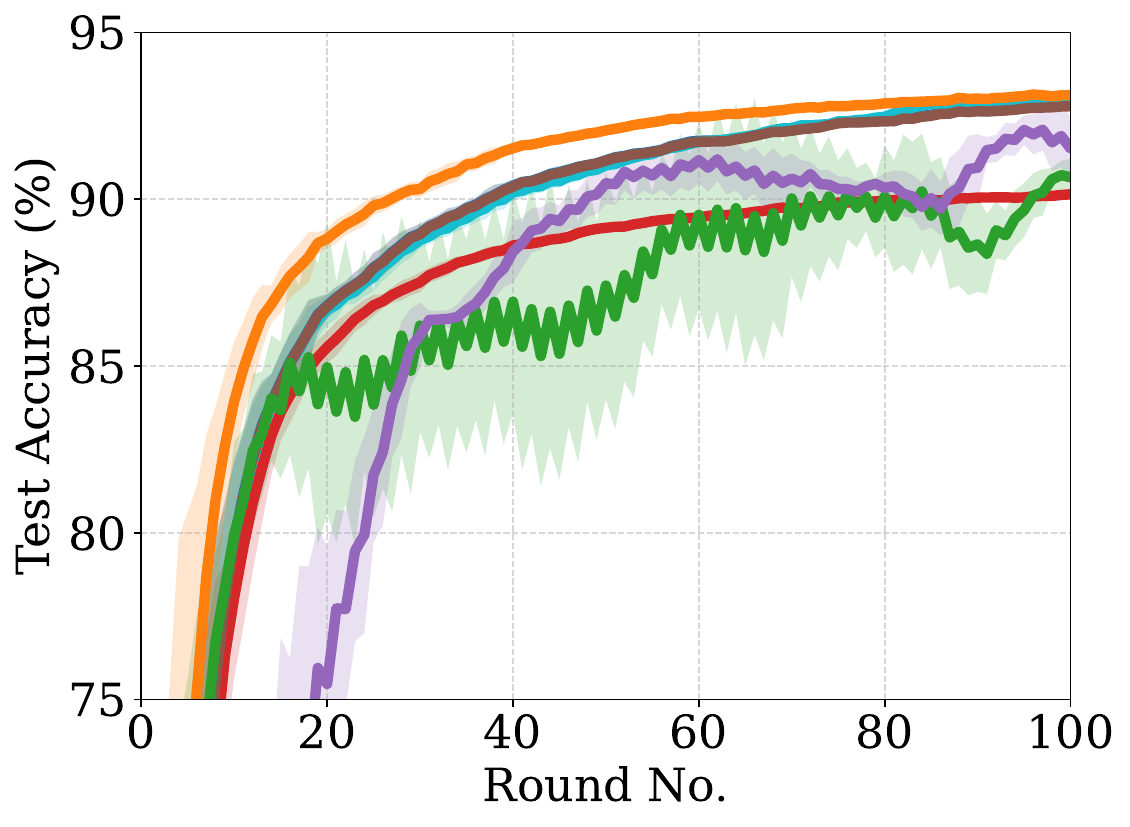}
    \caption{$20$ clients}
  \end{subfigure}
  \begin{subfigure}[b]{0.32\linewidth}
    \includegraphics[width=\linewidth]{figures/svhn_acc_plot.pdf}
    \caption{$50$ clients}
  \end{subfigure}
  \begin{subfigure}[b]{0.32\linewidth}
    \includegraphics[width=\linewidth]{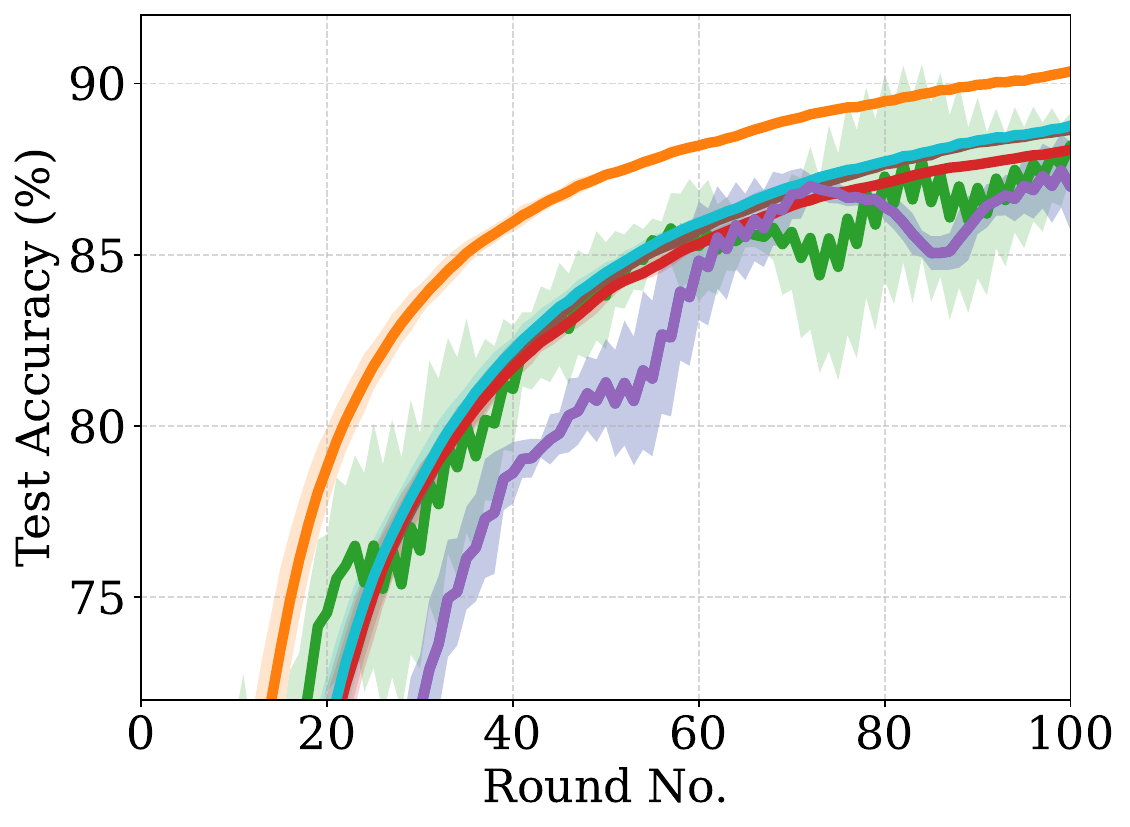}
    \caption{$100$ clients}
  \end{subfigure}
  \caption{Effect of number of clients on the performance of \texttt{FedRPCA} and baselines. Improvement offered by \texttt{FedRPCA} increases as the number of clients increases. }
  \label{fig:client_plots}
\end{figure}
\begin{figure}[!h]
  \centering
  {
  \includegraphics[height=0.34cm]{figures/common_legend.pdf}}   
  \begin{subfigure}[b]{0.32\linewidth}
    \includegraphics[width=\linewidth]{figures/svhn_acc_plot.pdf}
    \caption{Rank = $4$}
  \end{subfigure}
  \begin{subfigure}[b]{0.32\linewidth}
    \includegraphics[width=\linewidth]{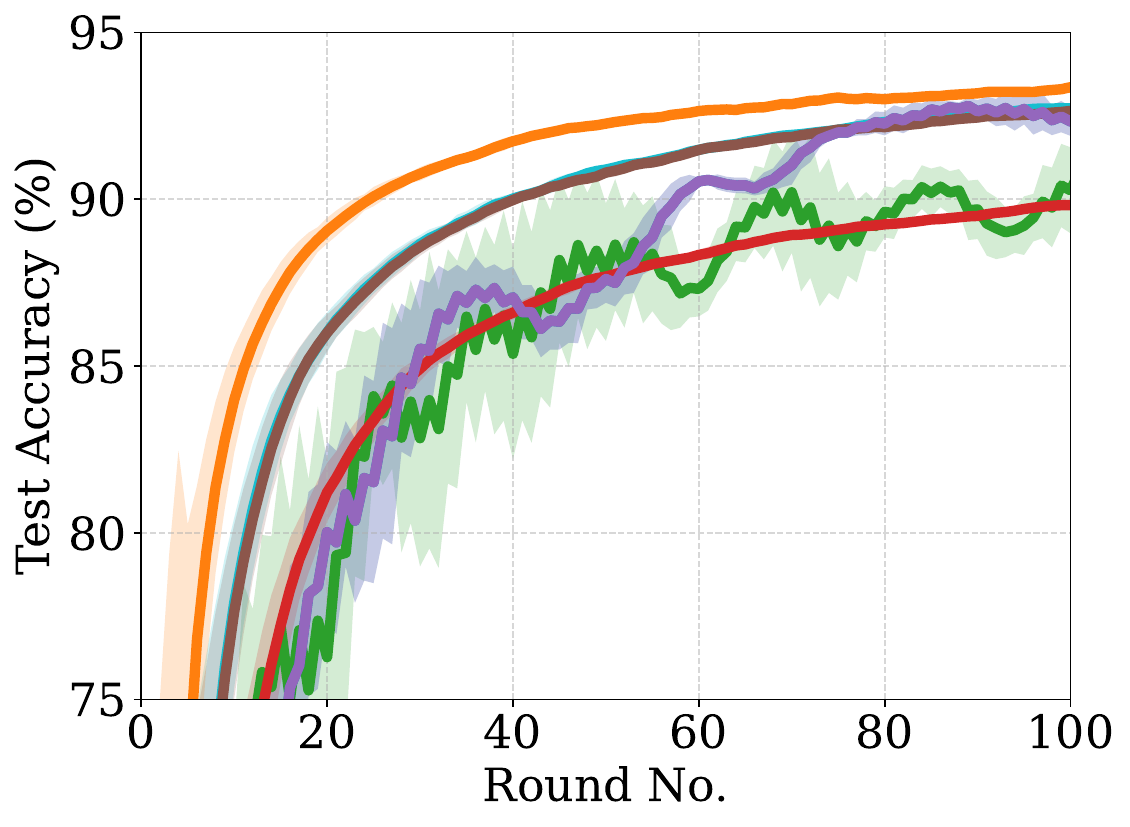}
    \caption{Rank = $8$}
  \end{subfigure}
  \begin{subfigure}[b]{0.32\linewidth}
    \includegraphics[width=\linewidth]{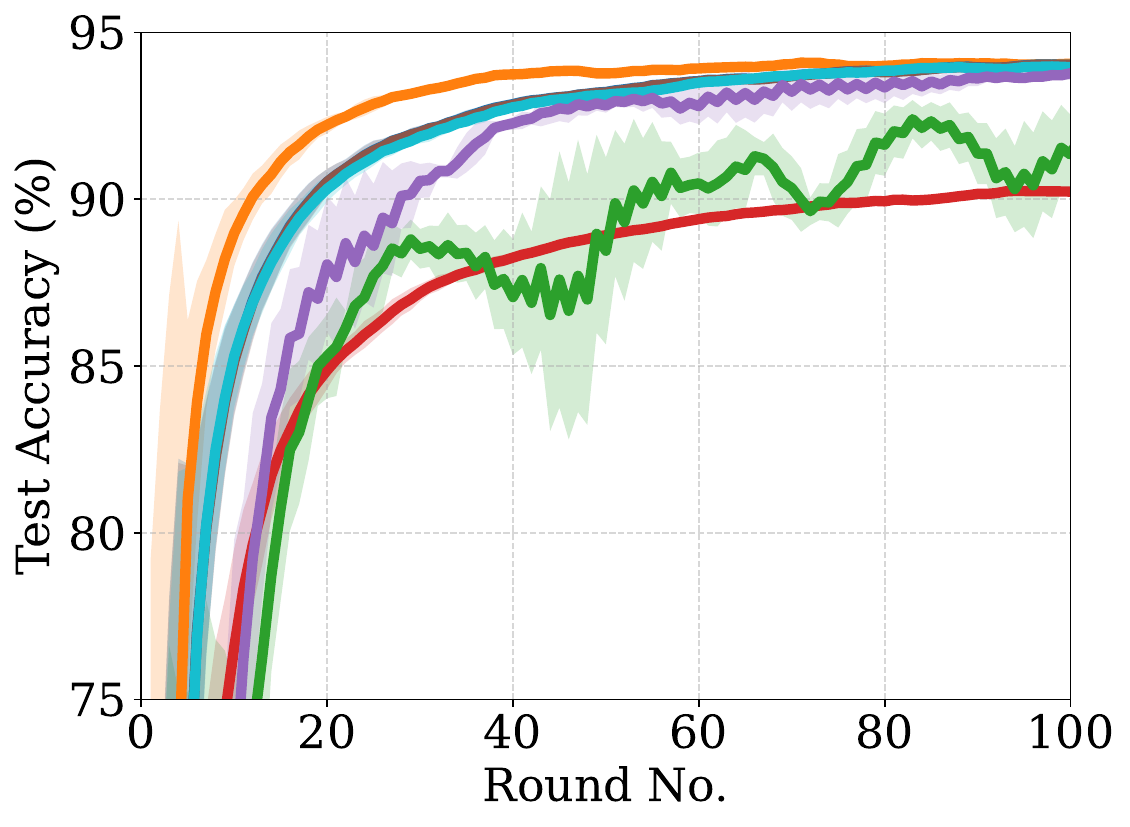}
    \caption{Rank = $32$}
  \end{subfigure}
  \caption{Effect of \texttt{LoRA} fine-tuning rank on the performance of \texttt{FedRPCA} and baselines. As the rank increases, the accuracy gap between \texttt{FedRPCA} and other baselines decreases; nonetheless \texttt{FedRPCA} still converges faster.}
  \label{fig:rank_plots}
\end{figure}


\subsection{Ablation Results on 20News Dataset}
\label{appendix:ablation_20News}
Here we extend our original ablation experiments conducted on the SVHN dataset to the 20News dataset. We omit comparisons with \texttt{SCAFFOLD} in this setting, as it did not perform competitively relative to other baselines. Overall, we observe similar trends: the performance gains from \texttt{FedRPCA} increase with greater client heterogeneity and a larger number of clients, while the speed-up remains consistent across different $\texttt{LoRA}$ fine-tuning ranks.

\begin{figure}[!h]
  \centering
  {
  \includegraphics[height=0.34cm]{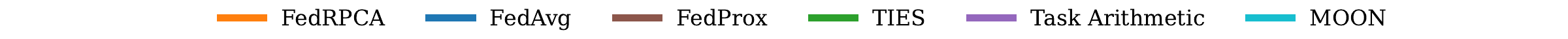}}
  \begin{subfigure}[b]{0.32\linewidth}
    \includegraphics[width=\linewidth]{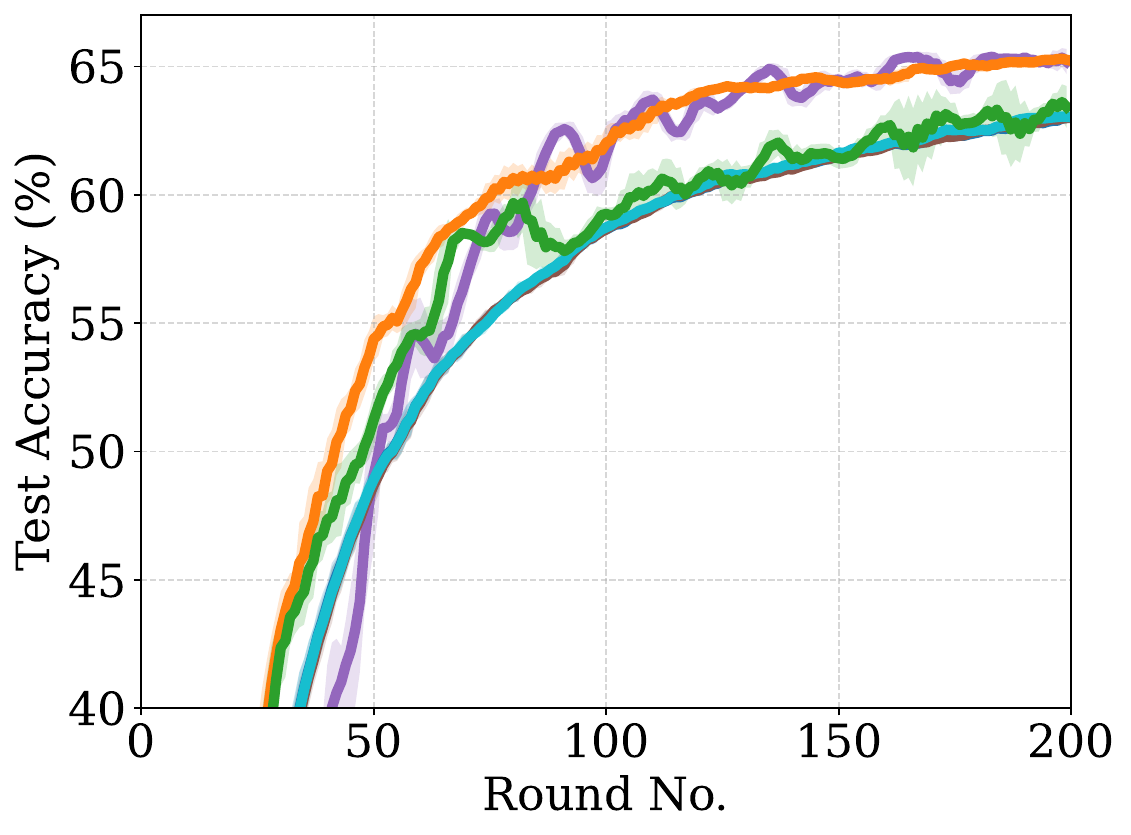}
    \caption{$\alpha = 10$}
  \end{subfigure}
  \begin{subfigure}[b]{0.32\linewidth}
    \includegraphics[width=\linewidth]{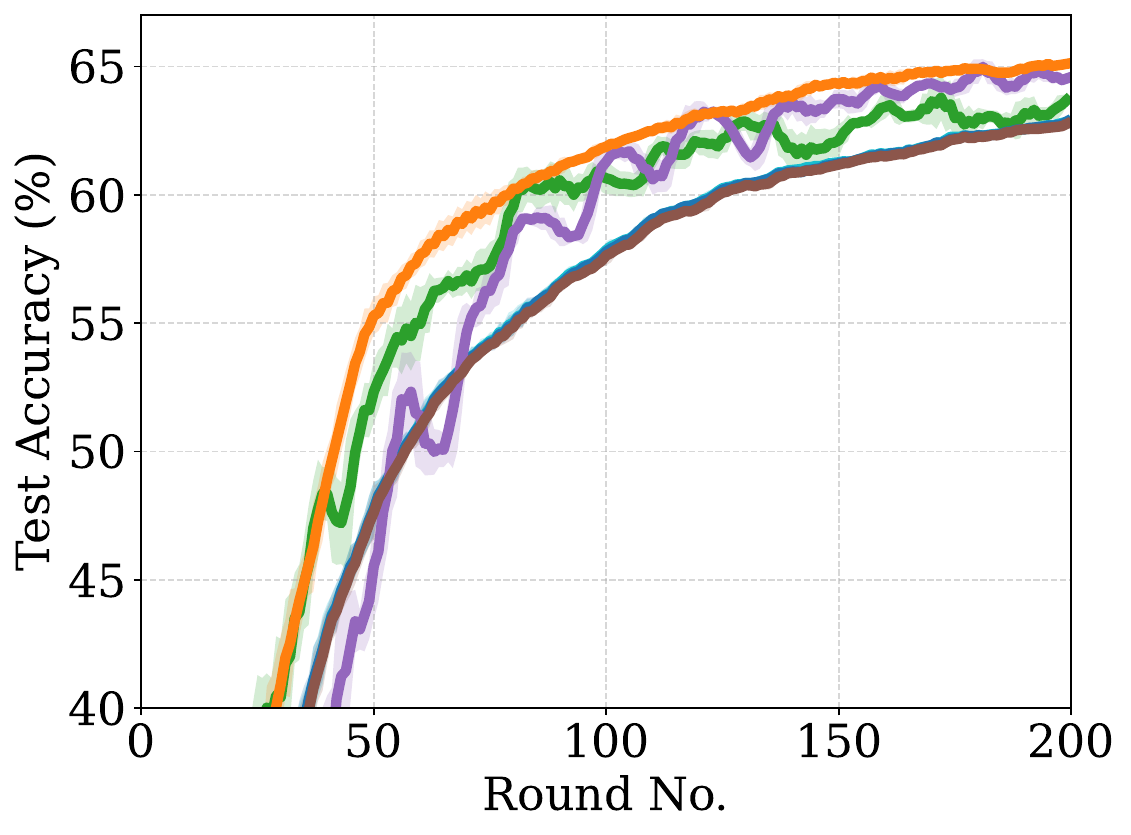}
    \caption{$\alpha = 1.0$}
  \end{subfigure}
  \begin{subfigure}[b]{0.32\linewidth}
    \includegraphics[width=\linewidth]{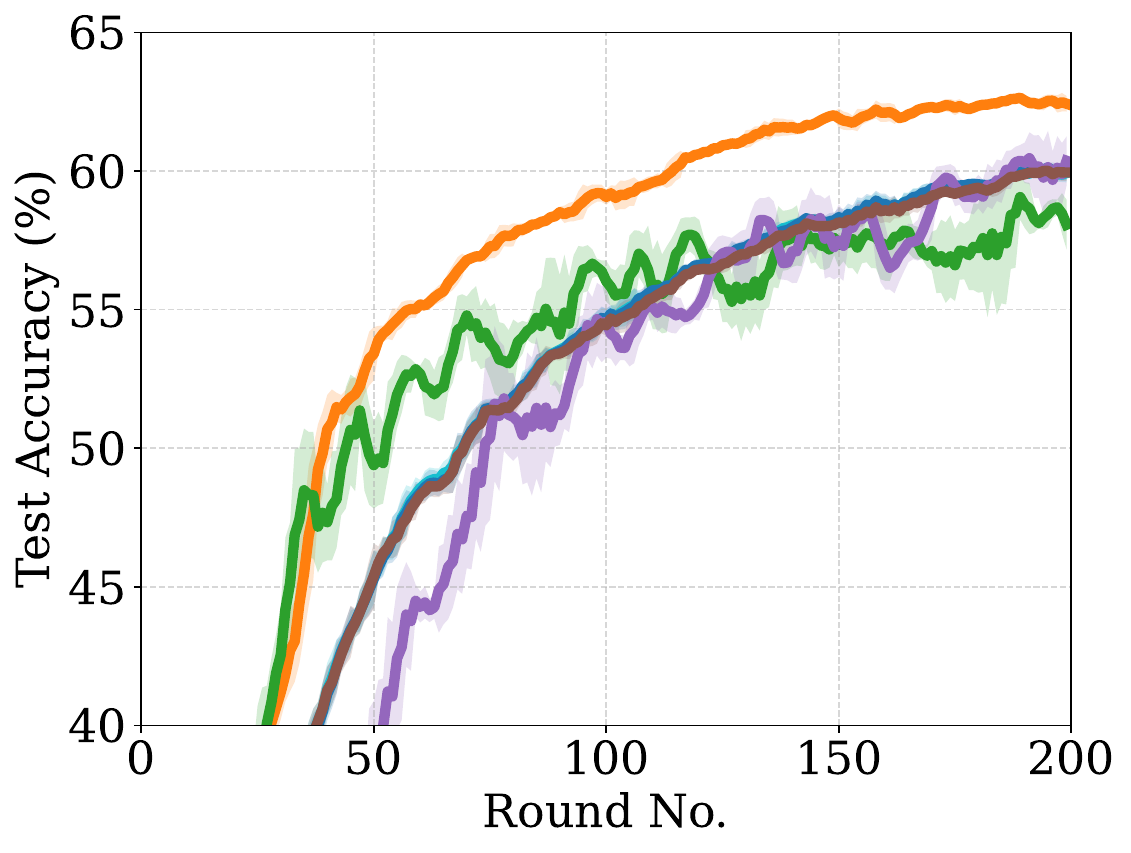}
    \caption{$\alpha = 0.1$}
  \end{subfigure}
  \caption{Effect of heterogeneity ($\alpha$) on the performance of \texttt{FedRPCA} and baselines when fine-tuning on 20News dataset.}
  \label{fig:hetero_plots}
\end{figure}

\begin{figure}[!h]
  \centering
  {
  \includegraphics[height=0.34cm]{figures/new_common_legend.pdf}}
  \begin{subfigure}[b]{0.32\linewidth}
    \includegraphics[width=\linewidth]{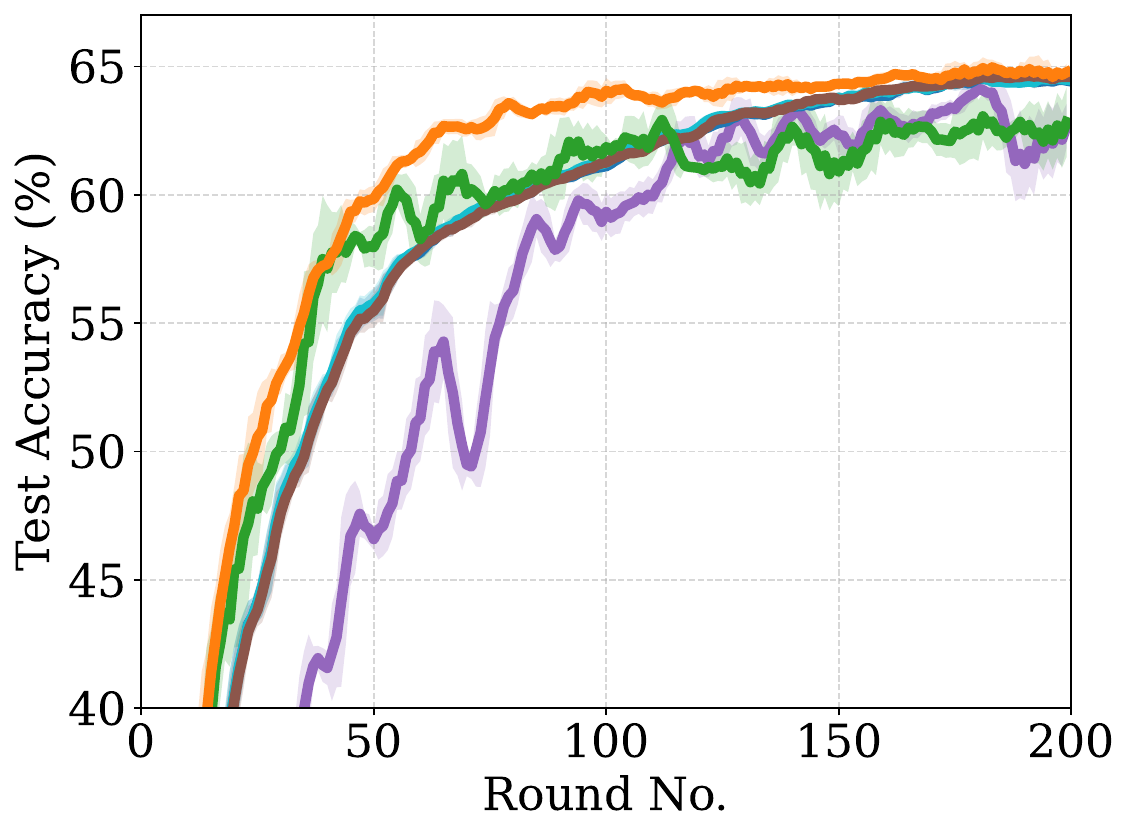}
    \caption{$20$ clients}
  \end{subfigure}
  \begin{subfigure}[b]{0.32\linewidth}
    \includegraphics[width=\linewidth]{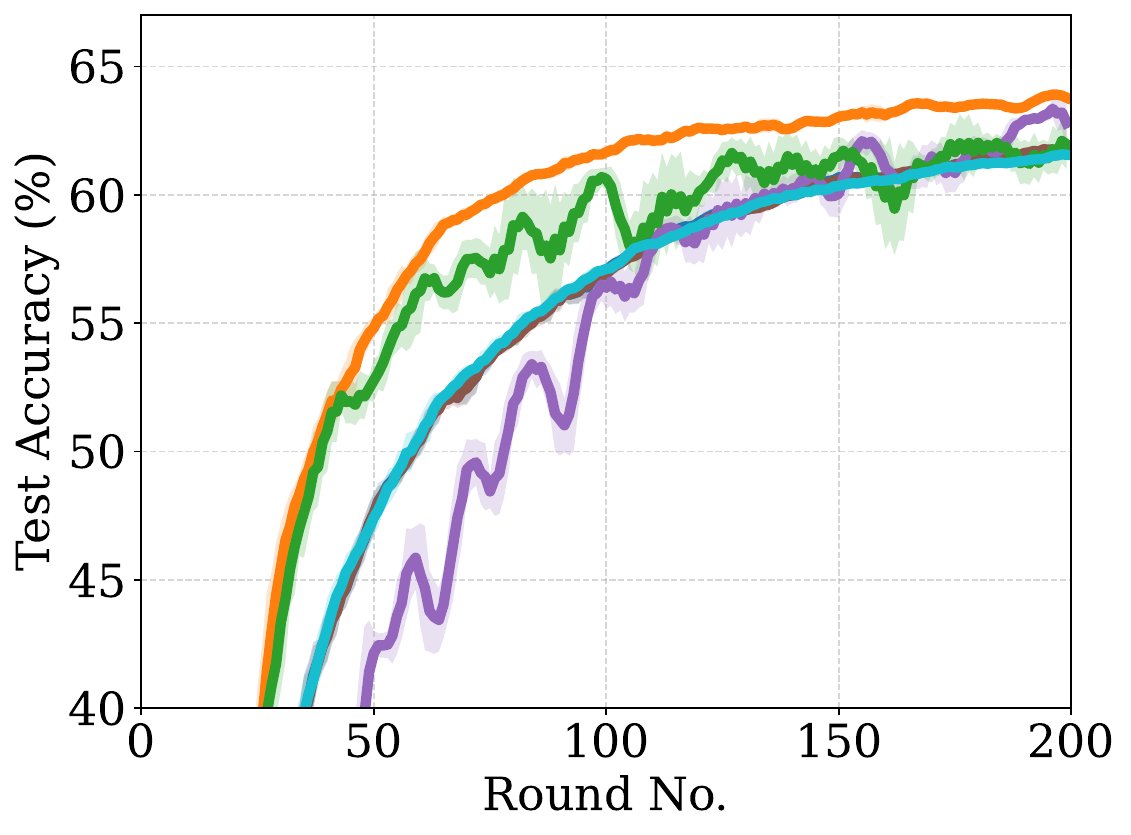}
    \caption{$50$ clients}
  \end{subfigure}
  \begin{subfigure}[b]{0.32\linewidth}
    \includegraphics[width=\linewidth]{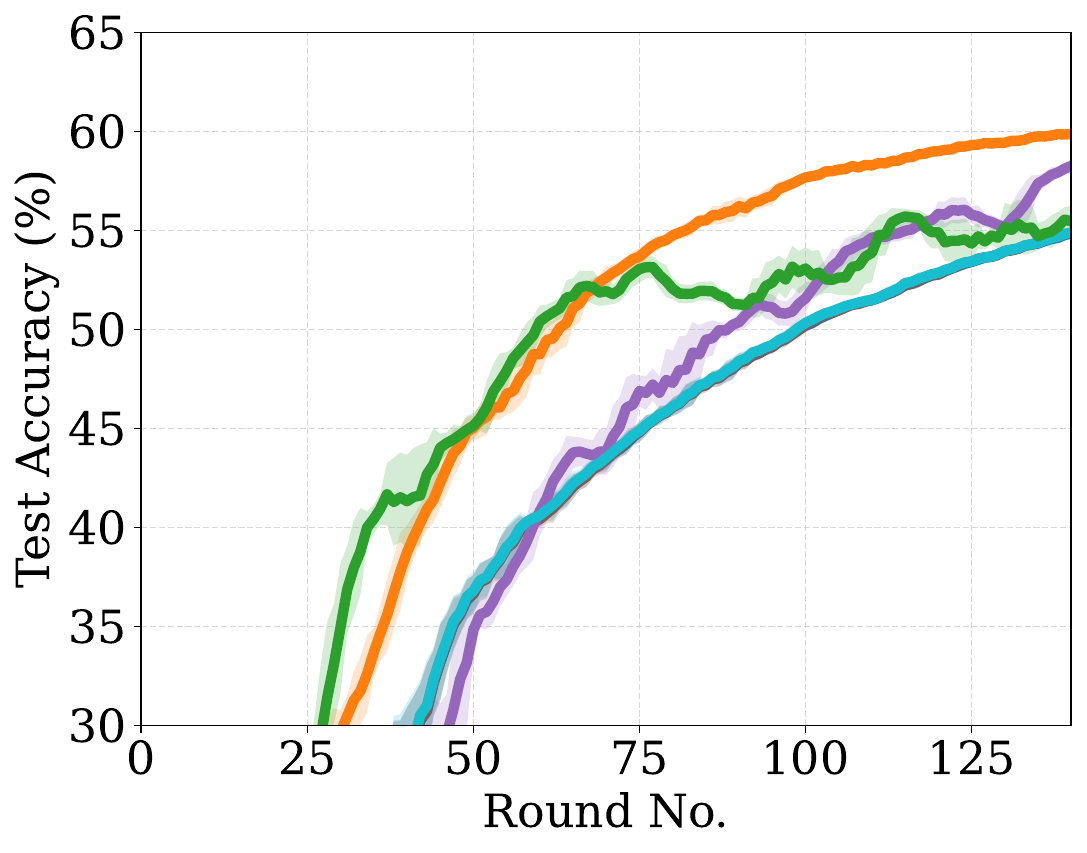}
    \caption{$100$ clients}
  \end{subfigure}
  \caption{Effect of number of clients on the performance of \texttt{FedRPCA} and baselines when fine-tuning on 20News dataset.}
  \label{fig:client_plots}
\end{figure}
\begin{figure}[!h]
  \centering
  {
  \includegraphics[height=0.34cm]{figures/new_common_legend.pdf}}        
  \begin{subfigure}[b]{0.32\linewidth}
    \includegraphics[width=\linewidth]{figures/20New_acc_plot.pdf}
    \caption{Rank = $4$}
  \end{subfigure}
  \begin{subfigure}[b]{0.32\linewidth}
    \includegraphics[width=\linewidth]{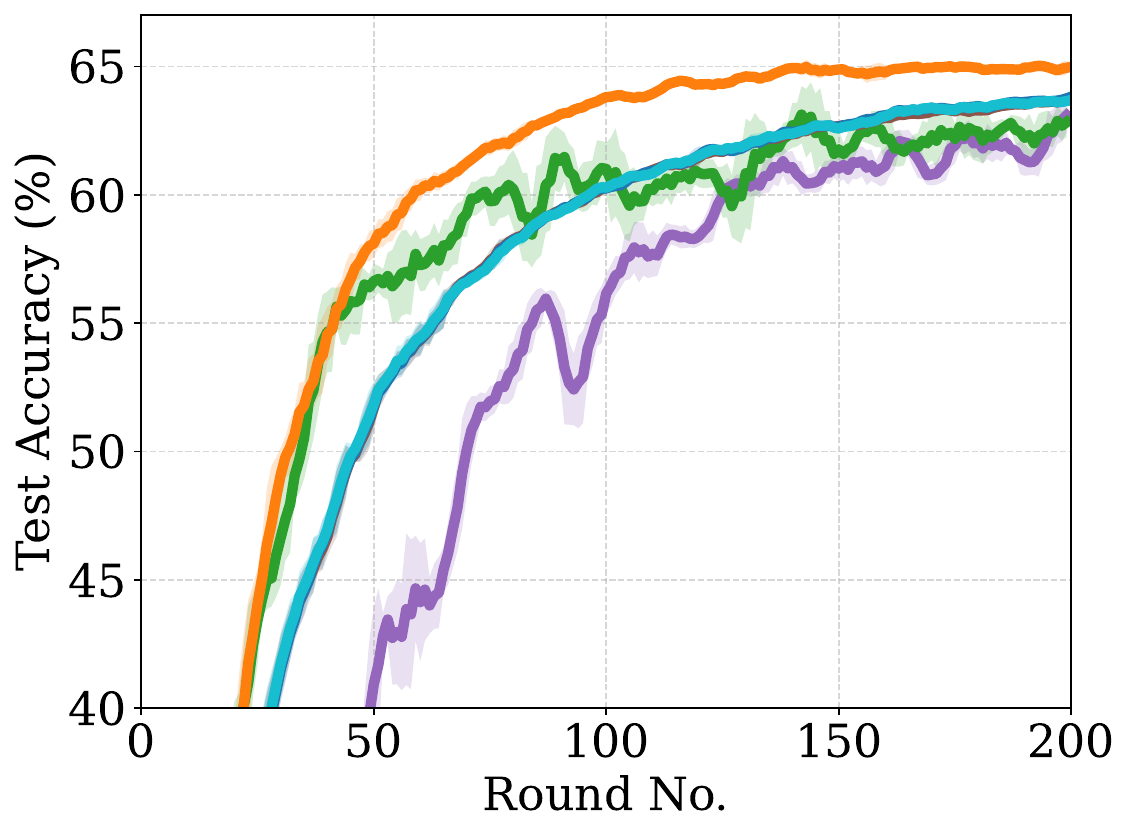}
    \caption{Rank = $8$}
  \end{subfigure}
  \begin{subfigure}[b]{0.32\linewidth}
    \includegraphics[width=\linewidth]{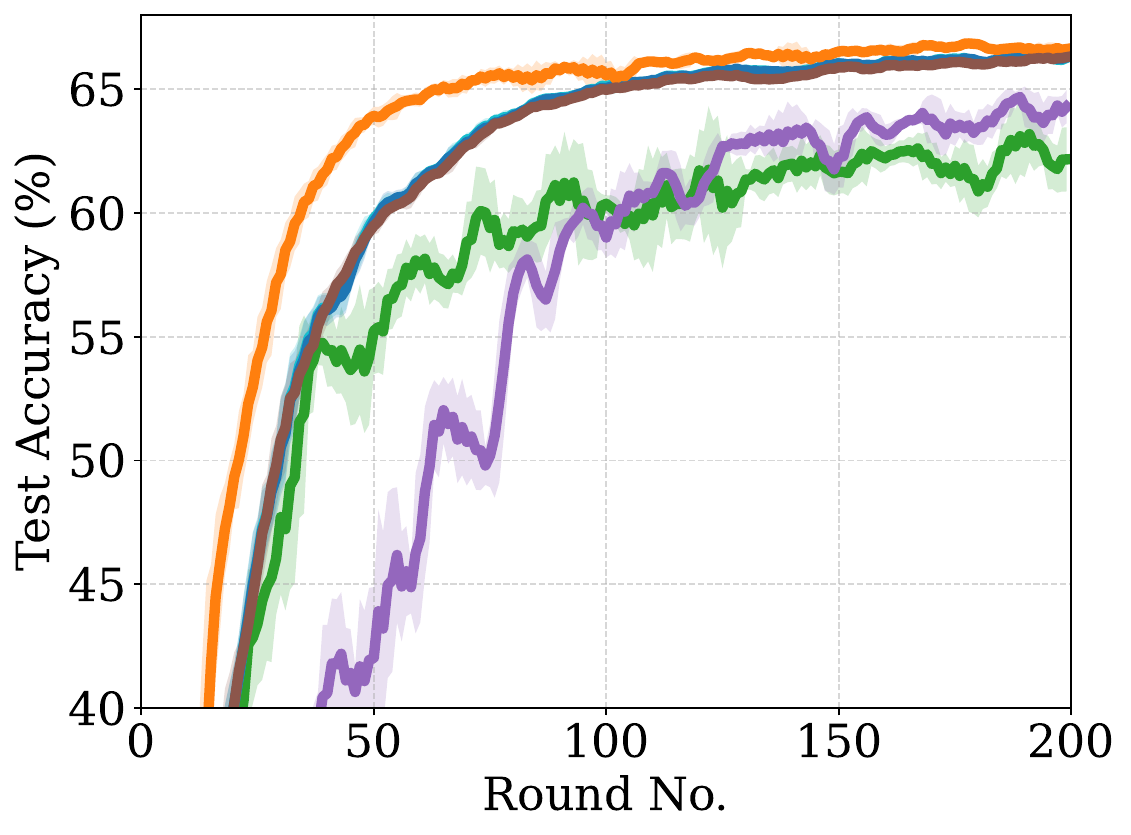}
    \caption{Rank = $32$}
  \end{subfigure}
  \caption{Effect of \texttt{LoRA} fine-tuning rank on the performance of \texttt{FedRPCA} and baselines when fine-tuning on 20News dataset.}
  \label{fig:rank_plots}
\end{figure}

\end{document}